\documentclass[10pt,twocolumn,letterpaper]{article}

\usepackage[pagenumbers]{cvpr}

\makeatletter
\@namedef{ver@everyshi.sty}{}
\makeatother
\usepackage{tikz}

\usepackage{url}
\usepackage{caption}
\usepackage{subcaption}
\usepackage{comment}
\usepackage[utf8]{inputenc} 
\usepackage[T1]{fontenc}    
\usepackage{amsfonts}       
\usepackage{nicefrac}       
\usepackage{microtype}      
\usepackage{xcolor}         

\usepackage{graphicx}
\usepackage{amsmath}
\usepackage{amssymb}
\usepackage{booktabs}
\usepackage{mmstyle}
\usepackage{multirow}
\usepackage{slashbox,pict2e}
\usepackage{float}
\usepackage[accsupp]{axessibility}  

\usepackage{times}
\usepackage{epsfig}
\usepackage{bm}
\usepackage{color}
\usepackage{colortbl}
\usepackage{pgfplots}
\usepackage[misc]{ifsym}
\usepackage{lipsum}

\usepackage{algorithm}
\usepackage[noend]{algpseudocode}

\usepackage[referable]{threeparttablex}

\usepackage{bm}
\usepackage{pifont}
\usepackage{multirow}

\usepackage{wrapfig}
\usepackage{url}
\usepackage{amsmath}
\usepackage{wrapfig}
\usepackage{float}

\definecolor{green}{RGB}{10, 179, 33}

\newlength\savewidth\newcommand\shline{\noalign{\global\savewidth\arrayrulewidth
		\global\arrayrulewidth 1pt}\hline\noalign{\global\arrayrulewidth\savewidth}}

\usepackage[pagebackref,breaklinks,colorlinks]{hyperref}

\usepackage[capitalize]{cleveref}
\crefname{section}{Sec.}{Secs.}
\Crefname{section}{Section}{Sections}
\Crefname{table}{Table}{Tables}
\crefname{table}{Tab.}{Tabs.}

\def\cvprPaperID{8663} 
\def\confName{CVPR}
\def\confYear{2023}

\begin{document}

\title{VLG: General Video Recognition with Web Textual Knowledge}

\author{Jintao Lin$^{1}$  \quad  \quad Zhaoyang Liu$^{2}$  \quad  \quad 
Wenhai Wang$^{3}$   \quad  \quad Wayne Wu$^{2}$  \quad  \quad Limin Wang\textsuperscript{${1,3}$ \Letter}\\
$^{1}$ State Key Laboratory for Novel Software Technology, Nanjing University, China\\
$^{2}$ SenseTime Research \ \ $^{3}$ Shanghai AI Laboratory\\
\tt\small jintaolin@smail.nju.edu.cn\ \ zyliumy@gmail.com\ \ wangwenhai@pjlab.org.cn\\
\tt\small wuwenyan0503@gmail.com\ \ lmwang@nju.edu.cn
}
\maketitle

\begin{abstract}
    Video recognition in an open and dynamic world is quite challenging, as we need to handle different settings such as close-set, long-tail, few-shot and open-set.
    By leveraging semantic knowledge from noisy text descriptions crawled from the Internet, we focus on the general video recognition (GVR) problem of solving different recognition tasks within a unified framework.
    The core contribution of this paper is twofold. First, we build a comprehensive video recognition benchmark of Kinetics-GVR, including four sub-task datasets to cover the mentioned settings.
    To facilitate the research of GVR, we propose to utilize external textual knowledge from the Internet and provide multi-source text descriptions for all action classes.
    Second, inspired by the flexibility of language representation, we present a unified visual-linguistic framework (VLG) to solve the problem of GVR by an effective two-stage training paradigm.
    Our VLG is first pre-trained on video and language datasets to learn a shared feature space, and then devises a flexible bi-modal attention head to collaborate high-level semantic concepts under different settings.
    Extensive results show that our VLG obtains the state-of-the-art performance under four settings. The superior performance demonstrates the effectiveness and generalization ability of our proposed framework.
    We hope our work makes a step towards the general video recognition and could serve as a baseline for future research.
    The code and models will be available at \url{https://github.com/MCG-NJU/VLG}.
\end{abstract}

\vspace{-1mm}
\section{Introduction}

Similar to image classification, the existing video recognition tasks are roughly grouped into four settings:
close-set~\cite{kay2017kinetics,carreira2019short,monfort2019moments,monfort2021multi}, long-tail~\cite{zhang2021videolt}, few-shot~\cite{zhu2018compound,zhu2020label,zhu2021closer} and open-set~\cite{acsintoae2021ubnormal,wang2021unidentified}, to mimic the realistic scenarios in practice.
With multiple video benchmarks~\cite{kay2017kinetics,soomro2012ucf101,goyal2017something,caba2015activitynet,carreira2019short},  
a number of works~\cite{wang2016temporal,tam,arnab2021vivit,zhang2021videolt,zhu2018compound,zhu2020label,bao2021evidential} have been developed to study video recognition in these diverse scenarios.

Though various video benchmarks and frameworks have been established in the last few years,
there still remain two problems: 
1) video datasets in different settings are normally collected from various data sources and naturally introduce domain bias. They are not suitable for studying general video representation. It is also inefficient for data organization and storage to use multiple benchmarks separately in different settings;
2) most works~\cite{feichtenhofer2016convolutional,carreira2017quo,tran2018closer,kumar2019protogan,shu2018odn} focus on addressing individual settings separately with different frameworks.
These separate investigations would ignore the potential sharing of knowledge among different settings. 
These problems severely impede the advance in video recognition as well as its application in the real world.
Accordingly, we aim to present a single video benchmark covering all these settings, and propose a simple framework to handle these different sub-problems under a unified perspective.

\begin{figure*}[t]
  \centering
\includegraphics[width=0.9\linewidth]{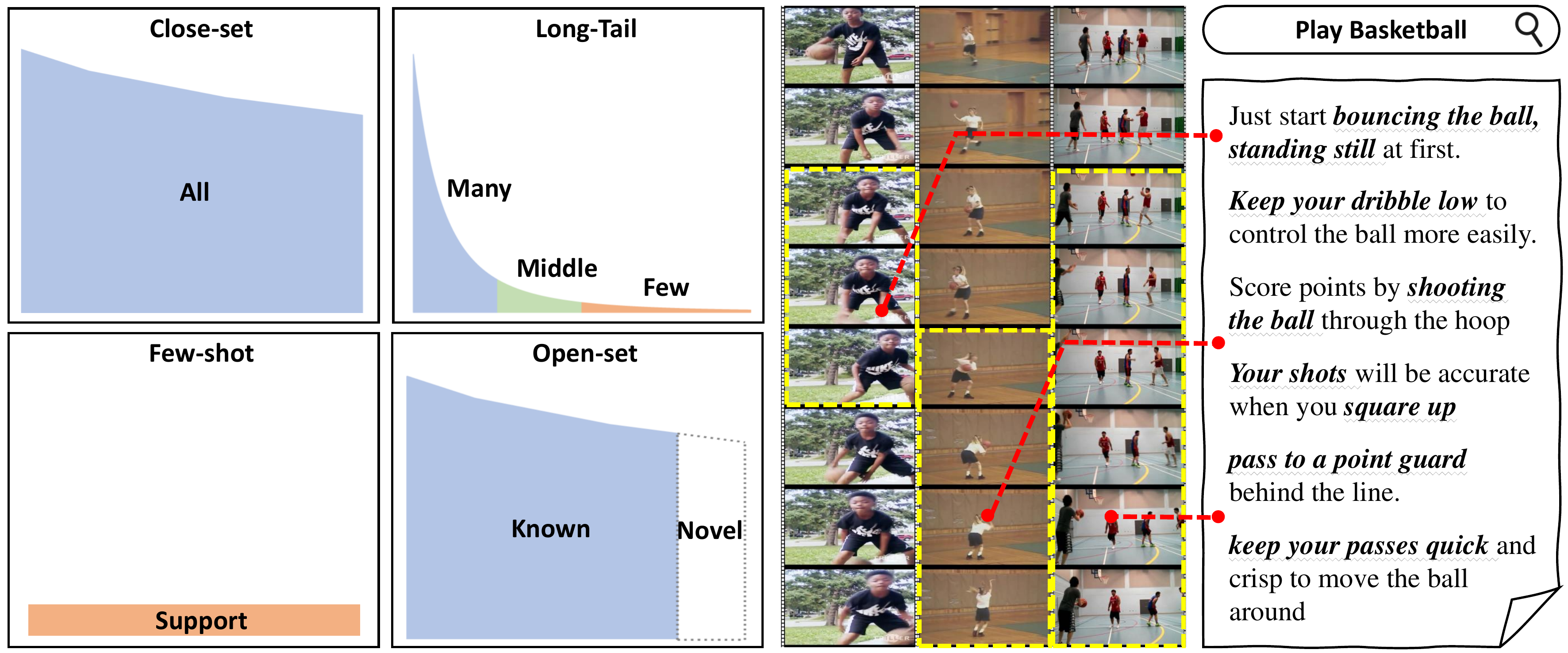}
\caption{
\textbf{Video label distribution of different scenarios and different modalities.}
As shown in the left, videos in GVR tasks have arbitrary distributions similar to natural data, such as close-set, long-tail, few-shot and open-set.
Most works only focus on coping with one aspect of them, while our method can use a unified framework
to address the GVR task by combining the advantages of video and text modalities.
The right part of the figure provides intuitive explanations for the correspondence between the videos and text modalities.
}
\vspace{-5pt}
\label{fig:moti}
\end{figure*}

\textit{To address the first problem:} we build a comprehensive video benchmark to dig into the \textbf{General Video Recognition} (GVR) problem, namely, covering video recognition under the following four settings.
As shown in the left of Figure~\ref{fig:moti}, 
this benchmark for GVR can cover a wide range of settings including close-set, long-tail, few-shot and open-set. Specifically, we curate a general video recognition benchmark {\bf Kinetics-GVR} from the Kinetics-400 dataset~\cite{carreira2017quo}, with four sub-settings:
Kinetics-Close, Kinetics-LT, Kinetics-Fewshot and Kinetics-Open, to mimic the video distribution of different scenarios in real-world applications. Our Kinetics-GVR aims to provide a solid benchmark to verify the performance of video recognition models under different distributions.

Since some works~\cite{radford2021learning,jia2021scaling,li2022mvp,yuan2021florence} have shown the efficacy of using natural language to supervise the visual representation learning, 
we intend to draw some extra knowledge (\textit{i.e.}, web text information) into our benchmark to facilitate the development of GVR. The extra web knowledge is expected to provide new cues for GVR. However, obtaining the paired text data for each video is prohibitively expensive.
As shown in the right of Figure \ref{fig:moti},
we observe that there are some connections between the video and text descriptions of its corresponding category. Specifically, the text descriptions for a specific video category exhibit some high-level semantic concepts to represent the static characteristics (\textit{e.g.}, scene) in space and dynamics (\textit{e.g.}, the steps to shooting) in time. 
In this sense, we hope that the text descriptions of video categories could also provide useful clues to learn a more general representation for GVR under different settings.
As a result, we also provide abundant text descriptions per-category in our benchmark to facilitate the research of GVR by crawling from the Internet.

\textit{To address the second problem:} 
we develop a unified framework to address general video recognition. Instead of dealing with each setting of video recognition with different frameworks, the unified framework would greatly reduce the work of hand-crafted design specific to each setting, and potentially increase its generalization ability due to the comprehensive consideration of all settings.

We find some recent visual-linguistic representation works, \textit{e.g.} CLIP~\cite{radford2021learning} and ALIGN~\cite{jia2021scaling},
can learn transferable visual models from natural language supervision, and show the promising performance on image recognition under different settings. 
However, there is still a lack of work to bridge the gap between video and text for general recognition under different scenarios.
Accordingly, we develop a video-language framework for general video recognition, termed as \textbf{VLG}.
VLG could benefit from the visual-linguistic models pretrained on the large-scale image-text pairs (\textit{e.g.}, CLIP~\cite{radford2021learning}), and connect video and text through customized temporal modeling. 
Our VLG leverages the rich semantic information of web text descriptions to guide the spatio-temporal feature learning.
Specifically, our method primarily contains four components:
1) The frame encoder to learn the visual representation for each frame.
2) The temporal module to model temporal features across frames for video domain adaption;
3) The language encoder to learn the textual representation for each sentence of category description.
4) The bi-modal attention head to perform general video recognition under different settings.
As text descriptions are directly collected from the Internet, they may include some noisy information. Thus we design a two-stage procedure to train our VLG: \textbf{Stage I} is to perform video-language pretraining,
adapting the encoders from the image domain to the video domain to learn a visual-linguistic representation.
\textbf{Stage II} filters out noisy texts and trains the bi-modal attention module to produce our final prediction. 
As demonstrated in experiments, our proposed VLG can effectively handle GVR under different settings of close-set, long-tail, few-shot, and open-set.

In summary, we make the following major contributions.
\begin{itemize}
    \item 1) We formulate the task of general video recognition (GVR) and establish a comprehensive benchmark to fairly test the performance of video recognition models under different data distributions. The benchmark for general video recognition based on Kinetics400~\cite{kay2017kinetics} comprises close-set, long-tail, few-shot and open-set, which shows different data distribution in practice.
    \item 2) To facilitate the research of GVR, we elaborately collect abundant text descriptions for each category. These extra textual knowledge exhibits more rich and high-level semantic concepts to represent the characteristics both in time and space, and contributes to the development of GVR.
    \item 3) We develop a unified video-language framework for general video recognition (VLG), which leverages the extensive web textual knowledge to effectively handle GVR under our customized two-stage learning strategy
    \item 4) Extensive experiments demonstrate the effectiveness of our VLG on the Kinetics-GVR for general video recognition under four settings.
\end{itemize}

\section{Related Work}
\noindent\textbf{Video Representation.}
Video recognition has made rapid progress from the early hand-craft descriptors~\cite{{klaser2008spatio,wang2013dense}} to current deep networks.
Deep neural networks can capture more general spatio-temporal representation from early two-stream networks, 3D-CNNs, and light-weight temporal modules to current transformer-based networks.
Two-stream networks~\cite{simonyan2014two,wang2016temporal,feichtenhofer2016convolutional} used two inputs of RGB and optical flow 
to separately model appearance and motion information in videos with a late fusion.
3D-CNNs~\cite{carreira2017quo,diba2018spatio,feichtenhofer2019slowfast,stroud2020d3d} proposed 3D convolution and pooling to model space and time jointly.
Light-weight temporal modules~\cite{tran2018closer,xie2018rethinking,zhou2018mict,jiang2019stm,kumawat2021depthwise,li2020tea,tam} were designed as powerful plugins to 
achieve the trade-off between efficacy and efficiency.
Recently, several works~\cite{bertasius2021space,arnab2021vivit} try to employ and adapt strong vision transformers 
to encode the spatial and temporal features jointly.
The aforementioned methods mostly focus on addressing the video recognition problem
only using visual modality in a supervised way, 
while ignoring the potentiality of natural language.

\noindent\textbf{Visual-Textual Learning.}
Visual-Textual Pretraining has made great progress on several down-stream vision tasks.
~\cite{cpd20} learned powerful video representation from a large-scale video-text pairs, with a contrastive learning method of CPD.
~\cite{miech2020end} proposed a new learning loss to address misalignments inherent in narrated videos. ~\cite{akbari2021vatt} proposed a framework for learning multimodal representations for unlabeled data.
Recently, CLIP~\cite{radford2021learning} and ALIGN~\cite{jia2021scaling} adopted simple noisy contrastive learning 
to obtain visual-linguistic representation from large-scale image-text web data.
Aligned with the success of image-language learning, lots of efforts have been made toward video-language learning~\cite{miech19endtoend,cpd2020,abs-2007-14937,wang2021actionclip,ju2021prompting}. \cite{cpd2020}, \cite{abs-2007-14937} and \cite{miech19endtoend} proposed to use text as supervision to learn video representations in a contrastive learning framework.
Some works also focus on a specific type of downstream tasks, \textit{e.g.} video-text VQA~\cite{wang2020general,kant2020spatially,singh2019towards}, video-text retrieval~\cite{dong2021dual,liu2021adaptive,yang2020tree}.
\cite{wang2021actionclip,ju2021prompting} adopted prompt engineering to reformulate their tasks into the same format as the pretraining objectives.
However, these methods cannot excavate the values of the noisy text descriptions data from the Internet,
leading to an unsatisfactory performance on real-world applications. To mitigate these issues,~\cite{tian2021vl} proposed to adopt class-wise text descriptions for long-tailed image recognition, while our method seeks to learn video-language representations and further extends the framework to varied video recognition settings.

\noindent\textbf{General Video Recognition (GVR).}
While GVR has not been defined in the existing literature,
we briefly summarize these sub tasks of GVR: long-tailed classification, few-shot learning and open-set classification.
Long-tailed classification has been extensively studied based on re-sampling data~\cite{buda2018systematic,chawla2002smote,chu2020feature,drumnond2003class,han2005borderline,shen2016relay}, 
re-weighting loss~\cite{cao2019learning,cui2019class,huang2016learning,khan2017cost,wang2017learning,tan2020equalization} 
and transferring strategy~\cite{kang2019decoupling,liu2019large,yin2019feature,zhou2020bbn,zhu2020inflated}.
Specifically,~\cite{zhang2021videolt} proposed to dynamically sample frames for long-tailed video recognition.
As for few-shot video classification, there has been a wide range of approaches,
including key frame representation memory~\cite{zhu2018compound}, 
adversarial video-level feature generation~\cite{kumar2019protogan}, 
and networks to utilize temporal information~\cite{bishay2019tarn,zhang2020few,cao2020few,zhu2021few,perrett2021temporal}.
As for open set video recognition, 
~\cite{shu2018odn} proposed ODN to gradually append new classes to the classification head,
~\cite{krishnan2018bar} adopted Bayesian deep learning to acknowledge unknown classes,
and~\cite{bao2021evidential} incorporated evidential learning for uncertainty-aware video recognition.

Compared with some current visual-linguistic approaches~\cite{wang2021actionclip,ju2021prompting} for tasks related to video recognition, our method can not only provide a comprehensive video-language representation to bridge the gap between videos and texts in different cases, but also effectively utilize noisy web text annotations in practical applications.

\begin{figure*}[t]
  \centering
\includegraphics[width=0.9\linewidth, ]{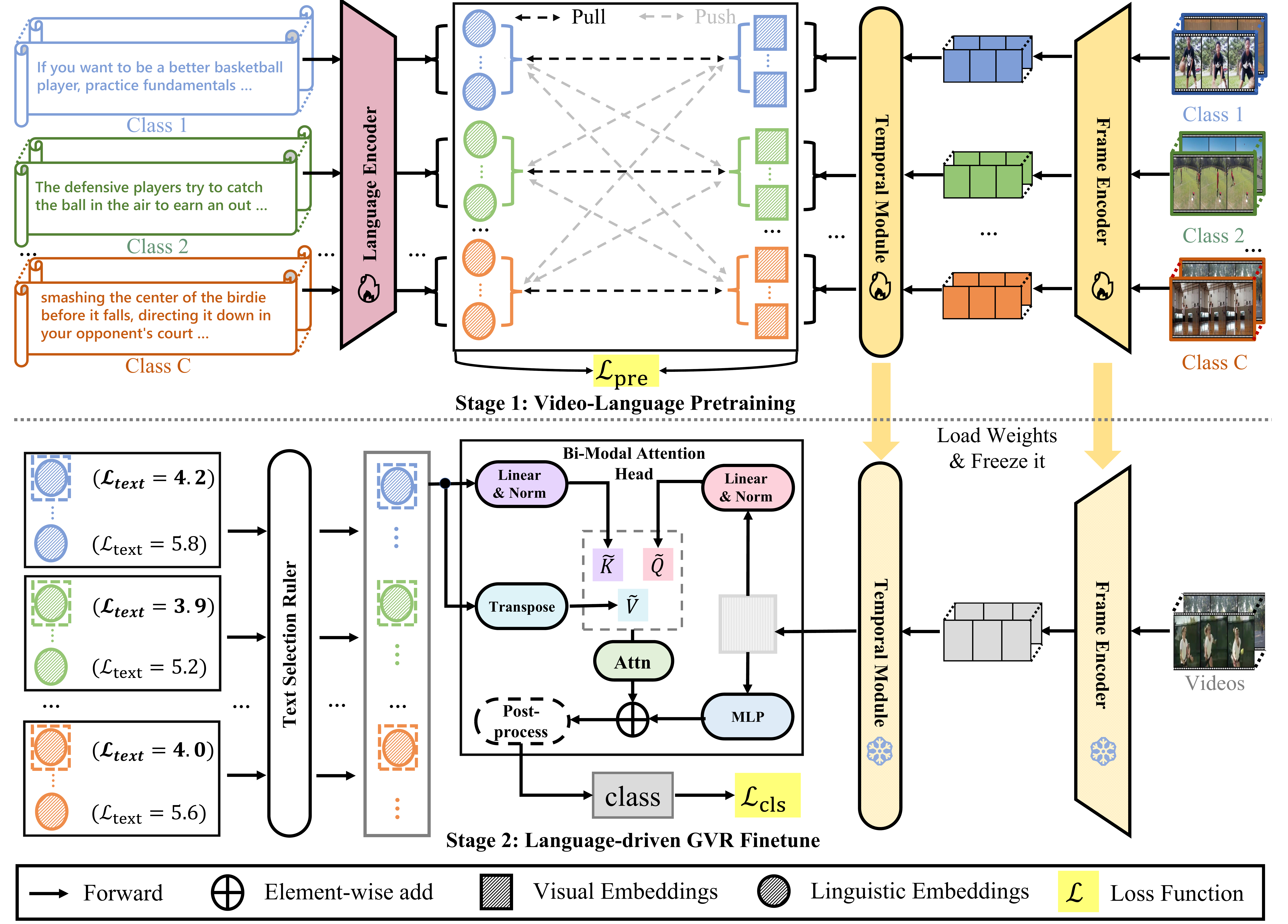}
\caption{
  \textbf{The pipeline of VLG.}
  The framework has two training stages.
  In the first stage, video-language pretraining (VLP) takes both the videos and text descriptions of each category
  as inputs, learning to link the two modalities through contrastive learning.
  In the second stage, embeddings of salient sentences, determined by the text selection ruler, are fed into the 
  bi-modal attention head to make final predictions.
}
  \vspace{-5pt}
\label{fig:pipeline}
\end{figure*}

\section{Kinetics-GVR}

To simulate the real-world video recognition from different scenarios,
we build a comprehensive video benchmark of Kinetics-GVR, consisting of Kinetics-Close, Kinetics-LT, Kinetics-Fewshot, and Kinetics-Open. 
Our benchmark is curated from the Kinetics-400~\cite{kay2017kinetics}. 
To obtain the text descriptions on labels, we crawl text entries from the Internet.

\noindent\textbf{Kinetics-Close.}
We adopt the original Kinetics400~\cite{kay2017kinetics} for close-set setting, 
which contains activities in daily life and has around 300k trimmed videos covering 400 categories.

\noindent\textbf{Kinetics-LT.}
For the long-tailed case, we construct the Kinetics-LT dataset,
which is a long-tailed version of Kinetics400 by sampling a subset following the Pareto distribution~\cite{liu2019large}.
Overall, it contains about 34.1K videos from 400 categories, 
with maximally 930 videos per class and minimally 5 videos per class.
The test set of it is the same as the original version.

\noindent\textbf{Kinetics-Fewshot.}
In the few-shot setting, we adopt the few-shot version of Kinetics~\cite{zhu2018compound}, 
which has been frequently used in previous works~\cite{zhu2020label,bishay2019tarn,perrett2021temporal}.
In this setup, 100 videos from 100 classes are selected, with 64/12/24 classes used for train/val/test.

\noindent\textbf{Kinetics-Open.}
For the open-set case, we split the Kinetics400 into two parts,
with 250 categories for training and the remaining 150 categories for evaluation.
Videos in the training set and validation set are from different categories.

\noindent\textbf{Text descriptions.}
The text descriptions are mainly crawled from Wikipedia~\cite{wiki} and wikiHow~\cite{wikihow}.
We first use the label as the keyword to search for the best matching entry.
Then, we filter out some irrelated parts of the entries, such as "references", and "bibliography", \textit{etc.},
to obtain the external text descriptions for each class.
In addition, we also append 96 prompt sentences for each class as basic descriptions,
which are generated by filling the pre-set templates, like `\texttt{a video of a \{label\}}', with label names.

\noindent For more details about these datasets, 
please refer to the Sec.~\ref{sec:benchmark} of Appendices.

\section{Method}

We first introduce the architecture of our proposed framework in Sec.~\ref{sec:method_architecture},
and then discuss its training strategy in Sec.~\ref{sec:method_optim}.
Finally, we present how to adapt our framework for different tasks in Sec.~\ref{sec:method_inference}.

\subsection{Overview}
\label{sec:method_architecture}

To effectively connect the video and language such that
language concepts can relate to visual representations for general video recognition,
we adopt a transformer-based Siamese network architecture~\cite{radford2021learning}, consisting of a video encoder ${\Phi}_\text{video}(\cdot)$
and a language encoder ${\Phi}_\text{text}(\cdot)$, to provide the visual representation and linguistic representation respectively.
Specially, the video encoder ${\Phi}_\text{video}(\cdot)$ is constructed with a frame encoder ${\Phi}_\text{img}(\cdot)$ 
followed by a temporal module ${\Phi}_\text{temp}(\cdot)$, 
which aggregates spatial features obtained from ${\Phi}_\text{img}(\cdot)$ over the temporal dimension.

As shown in the top of the Figure~\ref{fig:pipeline},
we first randomly sample a batch of videos $\mathcal{V}=\left\{V_i \right\}_{i=1}^N$,
and the corresponding text sentences $\mathcal{T}=\left\{T_i \right\}_{i=i}^N$,
where $V_i$ and $T_i$ are of the same class, $N$ denotes the batch size, and each video contains $F$ frames $V=\left\{I_i \right\}_{i=1}^F$.
For texts $\mathcal{T}$, they are fed to the language encoder ${\Phi}_\text{text}(\cdot)$ to yield text embeddings ${E}^T$,
while for videos $\mathcal{V}$, they are fed to the video encoder ${\Phi}_\text{video}(\cdot)$ to yield video embeddings ${E}^V$,
by extracting frame features with ${\Phi}_\text{img}(\cdot)$ and then aggregating features along the temporal dimension with ${\Phi}_\text{temp}(\cdot)$:
\begin{equation}
    {E}_{i}^T = {\Phi}_\text{text}(T_{i}),
\end{equation}
\vspace{-20pt}
\begin{equation}
    {E}_{i}^V = {\Phi}_\text{video}(V_{i}) = 
    {\Phi}_\text{temp}(\left\{ {\Phi}_\text{img}(I_{1}),...,{\Phi}_\text{img}(I_{F})\right\}).
\end{equation}
After that, we use a bi-modal attention head to aggregate the visual and linguistic features and then obtain the final prediction, as shown in the bottom of Figure~\ref{fig:pipeline}.

As raw text descriptions crawled from the Internet are noisy, it is necessary to obtain the salient sentences (namely, clean text descriptions) described in Sec.~\ref{sec:method_optim}. The salient sentences reduce the impacts of noises for final prediction, 
which has been demonstrated in experiments in the Sec.~\ref{sec:abl_study} of the Appendix.
The bi-modal attention head dynamically fuses the video embeddings and text embeddings of salient sentences based on the attention weights. 
Specially, given video embedding $E^V \in \mathbb{R}^D$ and salient text embeddings of a certain class $E^T \in \mathbb{R}^{M\times D}$, we first calculate the query $\widetilde{Q} \in \mathbb{R}^D$, key $\widetilde{K} \in \mathbb{R}^{M\times D}$ and value $\widetilde{V} \in \mathbb{R}^{M\times D}$ of the attention operation.
\begin{gather}
    \widetilde{Q} = \text{Linear}(\text{LayerNorm}(E^V)), \\
    \widetilde{K} = \text{Linear}(\text{LayerNorm}(E^T)), \\
    \widetilde{V} = E^T,
\end{gather}
where $C$ is the class number, 
and $M$ is the maximum number of sentences for each class, 
corresponding to the number of sampled salient sentences.
Next, we adopt an attention operation to gather these $M$ salient sentence embeddings for $\widetilde{G}\in \mathbb{R}^{D}$:
\begin{equation}
    \widetilde{G} = \text{Softmax}(\frac{\widetilde{Q} {\widetilde{K}}^\mathsf{T}}{\sqrt{D}}) \widetilde{V}.
\end{equation}
Then, we perform broadcasting to gather the salient sentences embeddings over all the classes for $G\in \mathbb{R}^{C \times D}$, where $C$ is the class number. The final classification probabilities are obtained based on the video embeddings $E^V$ and enhanced text embeddings $G$:
\begin{gather}
    P^V = \text{Softmax} (\text{MLP}(E^V)), ~\label{eq:PV} \\
    P^T = \text{Softmax} (\text{sim}(E^V, G) / \tau), ~\label{eq:PT} \\
    P = P^V + P^T, \label{eq:two_terms}
\end{gather}

where $P$ is the classification probability of the video, consisting of two terms, respectively for classification probability based on video representation $P^V$, and classification probability based on language representation $P^T$. $\text{sim}(\cdot,\cdot)$ denotes cosine similarity and $\tau$ is a learned parameter.

\subsection{Training}
\label{sec:method_optim}
We train our framework in two stages, 
namely Video-Language Pretraining (\textit{VLP}) and Language-driven GVR Finetune,
and design specific loss functions, \textit{i.e.} $\mathcal{L}_\text{pre}$ and $\mathcal{L}_\text{cls}$,
respectively for pretraining and classification.

\noindent\textbf{Stage I: Video-Language Pretraining.} 
We jointly optimize the language encoder and video encoder together with the temporal module.
The video features would be enclosed to their related category descriptions with higher similarity,
and pulled away from irrelated sentences. 
Specially, we use two contrastive learning NCE losses respectively for video embeddings ${E}^V$ and text embeddings ${E}^T$:
\begin{equation}
\small
\mathcal{L}_\text{text}\!=\!-\frac{1}{|\mathcal{V}^+_i|}\!\sum_{V_j\!\in\!\mathcal{V}^+_i}\!\text{log} \frac{\text{exp}(\text{sim}({E}_{j}^V, {E}_{i}^T) /\!\tau)} {\sum_{V_k\in \mathcal{V}} \text{exp}(\text{sim}({E}_{k}^V, {E}_{i}^T) / \tau)},
\end{equation}
\begin{equation}
\small
\mathcal{L}_\text{video}\!=\!-\frac{1}{|\mathcal{T}^+_i|}\!\sum_{T_j\!\in\!\mathcal{T}^+_i}\!\text{log} \frac{\text{exp}(\text{sim}({E}_{j}^T, {E}_{i}^V) /\!\tau)} {\sum_{T_k\in \mathcal{T}} \text{exp}(\text{sim}({E}_{k}^T, {E}_{i}^V) / \tau)},
\end{equation}
where $\mathcal{L}_\text{video}$ and $\mathcal{L}_\text{text}$ represent the video and language losses respectively.
$\mathcal{V}^+_i$ indicates a subset of $\mathcal{V}$, where all videos are of the same category with the text $T_i$.
Similarly, all texts in $\mathcal{T}^+_i$ share the same class with the video $V_i$. 

To effectively promote our framework for learning to connect the cross-modal information with limited text corpus,
we adopt CLIP~\cite{radford2021learning} pretrained model as the teacher model to distill knowledge for better visual-linguistic representation.
To aggregate the frame features along the temporal dimension, the teacher model replaces the temporal module with the average pooling and outputs the same dimensions of embeddings as the student model.
Their visual-linguistic similarities are used as soft targets for training weights associated with the student networks by the following objective:
\begin{equation}
    S_{\mathcal{V}} = \frac{\text{exp}(\text{sim}({E}_{i}^V, {E}_{i}^T) / \tau)}{\sum_{V_j\in \mathcal{V}} \text{exp}(\text{sim}({E}_{j}^V, {E}_{i}^T) / \tau)},\ \ \ \ 
\end{equation}
\begin{equation}
    S_{\mathcal{T}} = \frac{\text{exp}(\text{sim}({E}_{i}^T, {E}_{i}^V) / \tau)}{\sum_{T_j\in \mathcal{T}} \text{exp}(\text{sim}({E}_{j}^T, {E}_{i}^V) / \tau)},
\end{equation}
\begin{equation}
    \mathcal{L}_\text{dist} = -S_{\mathcal{V}}' \cdot \text{log}S_{\mathcal{V}} - S_{\mathcal{T}}' \cdot \text{log}S_{\mathcal{T}},
\end{equation}
where $S$ and $S'$ are cosine similarity scores respectively produced by our model and the frozen CLIP model.
With this pretraining stage, our framework can not only learn great video-language representation,
but also reduce the risk of overfitting limited text corpus data. Therefore, we optimize the video encoder and language encoder via pretraining loss $\mathcal{L}_\text{pre}$,
 defined as a weighted sum of
$\mathcal{L}_\text{video}$, $\mathcal{L}_\text{text}$ and $\mathcal{L}_\text{dist}$:
\begin{equation}
    \mathcal{L}_\text{pre} = \alpha \cdot (\mathcal{L}_\text{video} + \mathcal{L}_\text{text}) + (1 - \alpha) \cdot \mathcal{L}_\text{dist}.
\end{equation}
Here, $\alpha$ is used to balance $\mathcal{L}_\text{video}$, $\mathcal{L}_\text{text}$ and $\mathcal{L}_\text{dist}$, which is set to 0.5 in our experiments.

\noindent\textbf{Stage II: Language-driven GVR Finetune.} In order to take advantage of the valid semantic information and video-language feature, the second stage aims to select the salient sentences by filtering out the noisy texts, and then finetune the bi-modal attention head with the ground truth label.

To filter out noisy texts, we design a training-free text selection ruler (\textit{TSR}) after obtaining the text embeddings, to sample the most discriminative sentences for each category.
Specially, we randomly choose $\lambda$ videos of each class to construct a video batch $V'$.
Then, we calculate $\mathcal{L}_\text{text}$ between each sentence and video batch $V'$. Finally, we select $M$ sentences with the smallest $\mathcal{L}_\text{text}$ for the following classification. Note that the TSR only needs to perform once at stage II.

To finetune the bi-modal attention head, we adopt two Cross Entropy losses $\mathcal{L}_\text{CE}$
for $P^V$ and $P^T$ (see Eq.~\ref{eq:PV} and Eq.~\ref{eq:PT}) respectively:

\begin{equation}
    \mathcal{L}_\text{cls} = \mathcal{L}_\text{CE}(P^V, \mathbf{y}) + \mathcal{L}_\text{CE}(P^T, \mathbf{y}),
\end{equation}
where $\mathbf{y}$ is the ground truth label.

\begin{table*}[t]
    \caption{
        \textbf{Results on Kinetics-Close.}
        By introducing the class-wise text descriptions,
        our model achieves superior performance to the existing approaches. 
        “IN” denotes ImageNet and "K400" denotes Kinetics400. 
        "-" indicates the numbers are not available for us. 
        "$\text{CLIP}^*$" denotes that the model is initialized with the weights pretrained on 400M image-text pairs provided in CLIP~\cite{radford2021learning}.
        The total GFLOPs are calculated by the number of views and GFLOPs (per-view).
    }
    \centering
    \resizebox{0.92\linewidth}{!}{
    \renewcommand\arraystretch{0.97}
    \setlength{\tabcolsep}{2.0mm}
    \begin{tabular}{l|c|c|c|c|c|c|c}
    \shline
    \small
    \renewcommand{\arraystretch}{0.1}
    \multirow{2}{*}{Method} & \multirow{2}{*}{Pretrain} & \multirow{2}{*}{Frame} & \multirow{2}{*}{Views} & \multirow{2}{*}{Top-1} & \multirow{2}{*}{Top-5} & GFLOPs & \multirow{2}{*}{Param (M)} \\
    &&&&&& (per-view) &
    \\
    \hline
    X3D-XL~\cite{feichtenhofer2020x3d} & \multirow{3}{*}{None} & - & $10 \times 3$ & 79.1 & 93.9 & 48.4 & 11.0 \\
    SlowFast, R101+NL~\cite{feichtenhofer2019slowfast} & - & 16 & $10 \times 3$ & 79.8 & 93.9 & 234.0 & 59.9\\
    MViT-B, 64$\times$3 ~\cite{fan2021multiscale} & & 64 & $3 \times 3$ & {\bf 81.2} & 95.1 & 455.0 & 36.6 \\
    \hline
    TSM, ResNeXt101~\cite{lin2019tsm} & \multirow{3}{*}{IN-1K} & 8 & $10 \times 3$ & 76.3 & - & - & -  \\
    TANet, R152~\cite{tam} &  & 16 & $4 \times 3$ & 79.3 & 94.1 & 242.0 & - \\
    TDN, R101~\cite{Wang_2021_CVPR} &  & 24 & $10 \times 3$ & {\bf 79.4} & 94.4 & 198.0 & 88.0 \\ 
    \hline
    ViViT-L/16x2~\cite{arnab2021vivit} & \multirow{5}{*}{IN-21K} & 32 & $4 \times 3$ & 80.6 & 94.7 & - & - \\
    TimeSformer-L~\cite{bertasius2021space} & & 8 & $1 \times 3$ & 80.7 & 94.7 & 2380.0 & 121.4 \\
    ViViT-L/16x2 (320)~\cite{arnab2021vivit} & & 32 & $4 \times 3$ & 81.3 & 94.7 & 3992.0 & 310.8 \\
    Swin-L (384)~\cite{liu2021video} & & 32 & $10 \times 5$ & 84.9 & 96.7 & 2107.0 & 200.0 \\
    MViTv2-L (312)~\cite{li2021improved} & & 40 & $5 \times 3$ & {\bf 86.1} & 97.0 & 2828.0 & 217.6 \\
    \hline
    ViViT-H/16x2~\cite{arnab2021vivit} & \multirow{2}{*}{JFT} & 32 & $4 \times 3$ & 84.8 & 95.8 & - & - \\
    TokenLearner 16at18 (L/10)~\cite{ryoo2021tokenlearner} & & - & $4 \times 3$ & {\bf 85.4} & 96.3 & 4076.0 & 450.0 \\
    \hline
    CLIP-Raw, R50~\cite{radford2021learning} & \multirow{5}{*}{$\text{CLIP}^*$} & 8 & $1 \times 1$ & 46.2 & 60.8 & 52.1 & 102.0 \\
    CLIP-Raw, ViT-B/16~\cite{radford2021learning} & ~ & 8 & $1 \times 1$ & 55.0 & 67.5 & 144.0 & 150.0 \\
    CLIP-Close, R50~\cite{radford2021learning} &  & 8 & $1 \times 1$ & 68.1 & 87.7 & 49.7 & 115.0 \\
    CLIP-Close, ViT-B/16~\cite{radford2021learning} & ~ & 8 & $1 \times 1$ & 78.9 & 93.5 & 141.0 & 106.0 \\
    ActionCLIP (ViT-B/16)~\cite{wang2021actionclip} & & 16 & $10 \times 3$ & {\bf 82.6} & 96.2 & 563.1 & 141.7 \\
    \hline
    VLG, R50 &  \multirow{6}{*}{$\text{CLIP}^*$} & 8 & $1 \times 1$ & 72.3 & 90.8 & 76.7 & 148.0 \\
    VLG, ViT-B/16 & ~ & 8 & $1 \times 1$ & 81.8 & 95.3 & 148.0 & 121.0 \\
    VLG, ViT-B/16 & ~ & 16 & $1 \times 1$ & 82.4 & 95.8 & 282.3 & 121.0 \\
    VLG, ViT-B/16 & ~ & 16 & $4 \times 3$ & 82.9 & 96.1 & 282.3 & 121.0 \\
    VLG, ViT-L/14 & ~ & 8 & $1 \times 1$ & 85.5 & 96.3 & 650.3 & 371.0 \\
    VLG, ViT-L/14 &  & 8 & $4 \times 3$ & \textbf{86.4} & \textbf{97.0} & 650.3 & 371.0\\
    \shline
\end{tabular}}
    \vspace{-5pt}
    \label{tbl:close_set}
  \end{table*}

\subsection{VLG for General Video Recognition}
~\label{sec:method_inference}
In most cases, given a query video and pre-selected text embeddings of salient sentences, we first feed the query video into the video encoder to obtain video embeddings. Then, the final result is predicted with the video embeddings and text embeddings of salient sentences through a video-language attention head. 
We follow this procedure in both the close-set setting and long-tailed setting.
For the few-shot setting, we use base videos to pretrain the encoders during the first stage. Then, we use support videos to select salient sentences when combining the linguistic features, or directly use the video embeddings from VLP for linear probe testing.
For the open-set setting, we follow the common procedure to train the framework, and insert a post-process step, which can be instantiated as the off-the-shelf open-set procedures (\textit{e.g.}, OpenMax~\cite{bendale2016towards}, Softmax with threshold, \textit{etc.}), to recognize the novel videos during inference.

\section{Experiments}

We first introduce the evaluation metrics for different settings in Sec.~\ref{sec:eval_metrics},
before presenting state-of-the-art results over all these four benchmarks: 
Kinetics-Close, Kinetics-LT, Kinetics-Fewshot, and Kinetics-Open,
respectively in Sec.~\ref{sec:exp_on_close}, Sec.~\ref{sec:exp_on_lt}, Sec.~\ref{sec:exp_on_few}, and Sec.~\ref{sec:exp_on_open}.
We then present some representative visualization in Sec.~\ref{sec:more}.
More details, such as \textbf{Experimental Settings} and \textbf{Ablation Studies}, \textit{etc.}, are provided respectively 
in Sec.~\ref{sec:implement_details} and Sec.~\ref{sec:abl_study} of supplementary material.

\subsection{Evaluation Metrics.}
\label{sec:eval_metrics}
We evaluate the performance of our framework under all of these four benchmarks.
Besides the top-1 classification accuracy over all classes,
for the \textit{long-tailed} setting, we also report the accuracy of three disjoint subsets:
\textit{many-shot classes} (more than 100 training videos in each class),
\textit{medium-shot classes} (20$\sim$100 training videos in each class),
\textit{few-shot classes} (less than 20 training videos in each class).
For the \textit{open-set} setting, 
we use \textit{F-measure} score as a balance between precision and recall.

\subsection{Experiments on Kinetics-Close}
\label{sec:exp_on_close}
In Table~\ref{tbl:close_set}, we compare our proposed methods with prior methods on Kinetice-Close, \textit{i.e.} Kinetics400.
There are mainly traditional CNN-based methods, Transformer-based methods and CLIP-based methods.
It can be seen that Transformer-based methods and CLIP-based methods achieve better performance than traditional methods.
Particularly, our models achieve higher accuracy than other competitors.
For example, our method achieves 82.9\% top-1 accuracy with ViT-B/16 frame encoder, 
which exceeds ActionCLIP, a CLIP-based method, with fewer video views.
For a fair comparison, we further test our 16 frame ViT-B/16 on the val list of ActionCLIP, and our VLG achieves a higher accuracy performance of 83.5\%. 
Moreover, when using ViT-L/14 as our visual backbone, 
our VLG can further achieve a higher accuracy of 86.4\%, with lower resolution (224px) and fewer computational costs than MViTv2-L (312).

To further demonstrate the superiority of the proposed VLG, we propose CLIP-Raw and CLIP-Close as our baselines on Kinetics-Close to make fair comparisons. 
CLIP-Raw directly adopts the original CLIP weights and model with only prompt sentences to validate the accuracy performance, while CLIP-Close removing language encoder consists of the frame encoder loading CLIP pretrained weights, temporal module and a linear classifier layer and is finetuned on Kinetics-Close for 100 epochs.
One can observe that our method also gets absolute accuracy gain against the baselines with ResNet-50 (72.3\% vs. 68.1\% vs. 46.2\%) and ViT-B/16 (81.8\% vs. 78.9\% vs. 55.0\%) backbones.
The results are desirable since our framework can take the advantages of the semantic information in the text descriptions.

\subsection{Experiments on Kinetics-LT}
\label{sec:exp_on_lt}
In Table~\ref{tbl:lt}, we can see that our VLG models are superior to conventional vision-based methods with the same video encoders.
Since there are few long-tailed methods specific to videos, 
we re-implement and report the performance of some representative image long-tailed methods on Kinetics400-LT,
such as $\tau$-normalized, cRT, NCM, LWS~\cite{kang2019decoupling}, PaCo~\cite{cui2021parametric}, and SSD-LT~\cite{li2021self},
which are all initialized with CLIP pretrained weights.
We also add an additional temporal pooling without introducing any new parameters to aggregate features along the temporal dimension for them.
In addition, we also build CLIP-LT and CLIP-Raw as our simple baseline based on CLIP to corroborate our method. CLIP-LT is built the same as CLIP-Close.

It can be seen that our proposed method is superior to prior visual-based methods with the same backbone.
For example, using the same ResNet-50 backbone, 
the overall accuracy of VLG reaches 60.8\%, which outperforms SSD-LT by 12.5 points (60.8\% vs. 48.3\%),
and 10.7\% better than PaCo (60.8\% vs. 50.1\%).
Moreover, when compared to CLIP baseline models,
the performance of our method is also promising, which is 7.4\% better than the CLIP-LT, and 14.6\% better than the CLIP-Raw (60.8\% vs. 53.4\% vs. 46.2\%).
When using ViT-B/16 as the backbone, the overall accuracy of VLG can further boost up to 70.7\%.

\subsection{Experiments on Kinetics-Fewshot}
\label{sec:exp_on_few}
Following~\cite{ju2021prompting}, we conduct two kinds of few-shot settings, \textit{i.e.},~\textbf{5-shot-5-way} and~\textbf{5-shot-C-way}.

\textbf{5-shot-5-way.} 
For a fair comparison, this setting adopts the publicly accessible few-shot splits.
During training, we simply use the base split for our first pretraining stage, without meta-learning paradigms.
During the evaluation, we report average results over 200 trials with random sampling on the test split.
Table~\ref{tbl:exp_on_few} presents the average top-1 accuracy, 
and our method clearly achieves significant performance.
Following CLIP~\cite{radford2021learning}, we directly adopt the linear probe to test the visual representation output from the video encoder,
which obtains 84.6\% top-1 accuracy and is higher than the traditional few-shot learning methods.
When combining the linguistic features, the performance can further boost up to 94.0\%.
We also use the same network settings with textual information following~\cite{ju2021prompting}, and achieve better performance.

\textbf{5-shot-C-way.}
We further investigate a more challenging experiment setting,
which samples 5 videos from the training set for each class as the base split,
and then directly evaluates the model on the standard Kinetics400 testing split.
For statistical stability,
we report the average results over 10 trials to ensure the reliability of results.
It can be seen that our model still obtains a superior performance (62.8\% vs. 58.5\%), which is also higher than~\cite{ju2021prompting}.

\begin{table}[t]
    \caption{
        \textbf{Results on Kinetics-LT.}
        Traditional Long-tailed methods use the same visual backbone.
        $\text{CLIP}^*$ denotes that the model is initialized by the CLIP~\cite{radford2021learning} weights.
        We report the overall accuracy and the accuracy of three disjoint subsets.
    }
    \centering
    \resizebox{1.\linewidth}{!}{
    \begin{tabular}{l|c|c|c|c|c|c}
    \shline
    \renewcommand{\arraystretch}{0.1}

    \multirow{2}{*}{Method} & \multirow{2}{*}{Pretrain} & \multirow{2}{*}{Backbone} & \multicolumn{4}{c}{Accuracy (\%)} \\
    \cline{4-7}
     & & & Overall & Many & Medium & Few  \\
    \shline
    TSN~\cite{wang2016temporal} & \multirow{4}{*}{ImageNet} & \multirow{4}{*}{ResNet-50} & 47.2 & 59.3 & 49.4 & 23.6 \\
    TSM~\cite{lin2019tsm} &  &  & 46.0 & 66.3 & 46.1 & 17.3 \\
    TANet~\cite{tam} &  & & 45.8 & 66.8 & 45.4 & 17.4 \\
    SlowOnly~\cite{feichtenhofer2019slowfast} &  & & 44.8 & 67.7 & 44.1 & 14.4 \\
    \hline
    NCM~\cite{kang2019decoupling} &  \multirow{6}{*}{$\text{CLIP}^*$} & \multirow{6}{*}{ResNet-50} & 41.8 & 53.0 & 42.3 & 24.6 \\
    cRT~\cite{kang2019decoupling} &   &  & 43.7 & 58.9 & 43.8 & 22.3 \\
    $\tau$-normalized~\cite{kang2019decoupling} & & & 43.9 & 63.8 & 43.1 & 18.5 \\
    LWS~\cite{kang2019decoupling} &  &  & 45.1 & 58.6 & 44.8 & 27.1 \\
    SSD-LT~\cite{li2021self} &  & & 48.3 & 59.6 & 49.1 & 30.0 \\
    PaCo~\cite{cui2021parametric} &  & & 50.1 & 60.1 & 50.3 & 35.8 \\
    \hline
    CLIP-Raw~\cite{radford2021learning} & \multirow{3}{*}{$\text{CLIP}^*$} & \multirow{3}{*}{ResNet-50} & 46.2 & 48.3 & 44.8 & 46.7 \\
    CLIP-LT~\cite{radford2021learning} & & & 53.4 & 70.3 & 53.3 & 31.1 \\
    VLG &  &  & 60.8 & 71.7 & 60.4 & 47.2 \\
    \hline
    CLIP-Raw~\cite{radford2021learning} & \multirow{3}{*}{$\text{CLIP}^*$} &
    \multirow{3}{*}{ViT-B/16} & 55.0 & 57.1 & 53.7 & 55.5 \\
    CLIP-LT~\cite{radford2021learning} & & & 63.8 & 79.7 & 63.8 & 42.8 \\
    VLG & &  & \textbf{70.7} & \textbf{81.9} & \textbf{69.7} & \textbf{58.3} \\
    \shline

\end{tabular}}
    \vspace{-3pt}
    \label{tbl:lt}
\end{table}

\begin{table}[t]
    \caption{
        \textbf{Results on Kinetics-Fewshot.}
        Here, $\mathcal{C}_{ALL}$ denotes the model is tested on all categories of the corresponding dataset,
        rather than only 5-way classification. 
        In Kinetics-fewshot, $\mathcal{C}_{ALL}=400$.
        "VLG-L" denotes our method with linear probe testing.
    }
    \centering
    \resizebox{0.93\linewidth}{!}{
    \renewcommand\arraystretch{0.97}
    \setlength{\tabcolsep}{2.0mm}
    \begin{tabular}{l|c|cc|c}
    \shline
    \renewcommand{\arraystretch}{0.1}

    Method & Backbone & K-shot & N-way & Top-1 \\
    \hline
    CMN~\cite{zhu2018compound} & \multirow{5}{*}{ResNet-50} & 5 & 5 & 78.9 \\
    TARN~\cite{bishay2019tarn} &  & 5 & 5 & 78.5 \\
    ARN~\cite{zhang2020few} & & 5 & 5 & 82.4 \\
    VLG-L & & 5 & 5 & 84.6 \\
    VLG & & 5 & 5 & \textbf{94.0} \\
    \hline
    E-Prompt~\cite{ju2021prompting} & \multirow{2}{*}{ViT-B/16} & 5 & 5 & 96.4 \\
    VLG & & 5 & 5 & \textbf{96.9} \\
    \hline
    E-Prompt~\cite{ju2021prompting} & \multirow{2}{*}{ViT-B/16} & 5 & $\mathcal{C}_{ALL}$ & 58.5 \\
    VLG & & 5 & $\mathcal{C}_{ALL}$ & \textbf{62.8} \\
    \hline
\end{tabular}}
    \label{tbl:exp_on_few}
    \vspace{-3pt}
\end{table}

\begin{table}[t]
	\caption{\textbf{Results on Kinetics-Open.}
    OLTR and VLG are both initialized by CLIP weight. With the same backbone, VLG outperforms OLTR among all thresholds.}
    \centering
    \label{tbl:open_set}
	\resizebox{1.\linewidth}{!}{
	\begin{tabular}{l|c|c|c|c|c|c}
    \shline
    \renewcommand{\arraystretch}{0.1}

    \multirow{2}{*}{Method} & \multirow{2}{*}{Post-process} & \multicolumn{5}{c}{F-measure} \\
    \cline{3-7}
     & & thr=0.1 & thr=0.2 & thr=0.3 & thr=0.5 & thr=0.7 \\
    \shline
    OLTR, R50~\cite{liu2019large} & Threshold & 0.490 & 0.504 & 0.513 & 0.502 & 0.458 \\
    VLG, R50 & Threshold & 0.610 & 0.639 & \textbf{0.654} & \textbf{0.610} & \textbf{0.469} \\
    VLG, R50 & OpenMax & \textbf{0.616} & \textbf{0.641} & 0.651 & 0.614 & 0.465 \\
    \shline
    VLG, ViT-B/16 & Threshold & 0.657 & 0.672 & 0.694 & \textbf{0.721} & 0.697 \\
    VLG, ViT-B/16 & OpenMax & 0.694 & 0.698 & \textbf{0.703} & 0.699 & 0.633 \\
    \shline
\end{tabular}
}
\vspace{-10pt}
\end{table}

\begin{figure*}[htb!]
    \centering
    \includegraphics[width=0.95\linewidth]{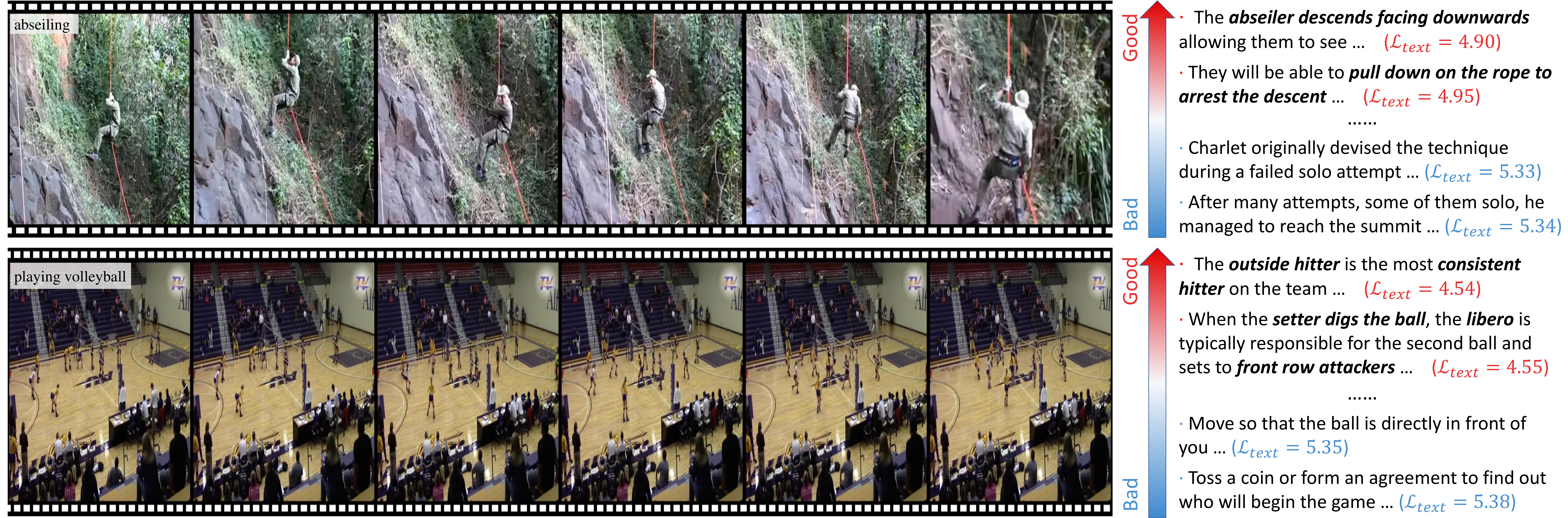}
    \caption{
        \textbf{Visualization of some text descriptions with corresponding $\mathcal{L}_\text{text}$}.
        The values of $\mathcal{L}_\text{text}$ reflect the saliency of these sentences,
        indicating the effectiveness of our TSR.
        It can be seen that sentences with words of specific concepts 
        (such as the "pull down on the rope" for "abseiling" and "libero" for "playing volleyball") 
        lead to small values of $\mathcal{L}_\text{text}$, indicating their saliency for classifying a certain category.
    }
    \label{fig:vis}
    \vspace{-10pt}
\end{figure*}

\subsection{Experiments on Kinetics-Open}
\label{sec:exp_on_open}
Openset video recognition aims to not only accurately classify known categories which have appeared in training,
but also recognize unknown categories which are not seen in training.
Without any other modifications to our framework,
we only adopt softmax with \textit{thresholds} and \textit{OpenMax}~\cite{bendale2016towards}
as a post-process on the prediction logits to obtain the classification results, as described in Sec.~\ref{sec:method_inference}.
In addition, we also re-implement the OLTR with CLIP initialization as a comparison.
As shown in Table~\ref{tbl:open_set}, we outperform OLTR~\cite{liu2019large} among all different threshold numbers, 
indicating the significance of our video-language representation.

\subsection{Qualitative Analysis on Salient Sentence}
\label{sec:more}
As shown in Figure~\ref{fig:vis}, we present some sentences of different categories sampled or filtered out by our TSR. 
We observe that our method can learn specific concepts or steps for each class, 
such as the "pull down on the rope" for "abseiling" and "libero" for "playing volleyball".
The salient sentences commonly contain these words of specific concepts in the category.
More examples of qualitative analysis on salient sentences are provided in the Appendix.

\subsection{More Results and Visualizations}
More results about ablation studies on the effectiveness of our video-language pretraining, 
CLIP pretrained weights, bi-modal attention head, temporal module, distillation loss and text selection ruler, \textit{etc.}, 
can be found in the Sec.~\ref{sec:abl_study} of supplementary material. 
Furthermore, we also provide the implementation details and evaluate the robustness of our method by altering the distribution of the labels.
In addition, we also provide additional visualization in the Appendix to illustrate the class-level performance improvement on long-tailed videos, 
examples of text descriptions, and more samples to show the relationship between videos and texts.

\section{Conclusions}

In this paper, we have studied the general video recognition (GVR) under four different settings. The GVR task enables us to examine the generalization ability of a video recognition model in real-world applications. To facilitate the research of GVR, we build comprehensive video benchmarks of Kinetics-GVR containing text descriptions for all action classes. Then, we propose a unified visual-linguistic framework (VLG) to accomplish the task of GVR. In particular, we present an effective two-stage training strategy to effectively adapt the image-text representation to video domain for GVR. Extensive results demonstrate that our VLG obtains the state-of-the-art performance under all settings on the Kinetics-GVR benchmark. We hope that the datasets and framework will help the future research in GVR.

{\small
\bibliographystyle{ieee_fullname}
\bibliography{egbib}
}

\appendix

\section*{Supplementary Material}

In this supplementary material, 
we first provide some details on video splits and text descriptions of our proposed benchmarks in Sec.~\ref{sec:benchmark}.
Then, we summarize the notations used in the paper in Sec.~\ref{sec:method_details}, 
and implementation details in Sec.~\ref{sec:implement_details}.
We also provide results of ablation studies and additional experiments respectively in Sec.~\ref{sec:abl_study} and Sec.~\ref{sec:additional_exp}.
Finally, we provide more visualization and discuss the limitation of our method, respectively in Sec.~\ref{sec:more_vis} and Sec.~\ref{sec:limitation}.
We also present the ethic statement and reproducibility statement of our method in Sec.~\ref{sec:negative} and Sec.~\ref{sec:reproduce} respectively.

\section{Benchmark Details}
\label{sec:benchmark}
\subsection{Kinetics-Close}
We directly adopt the original Kinetics400~\cite{kay2017kinetics} for close-set setting, 
which contains activities in daily life and has around 300k trimmed videos covering 400 categories.
Because of the expirations of some YouTube links,
some original videos are missing over time.
Our copy includes 240436 training videos and 19796 validation videos.

\begin{figure*}[t]
    \centering
    \resizebox{0.95\linewidth}{!}{
    \includegraphics{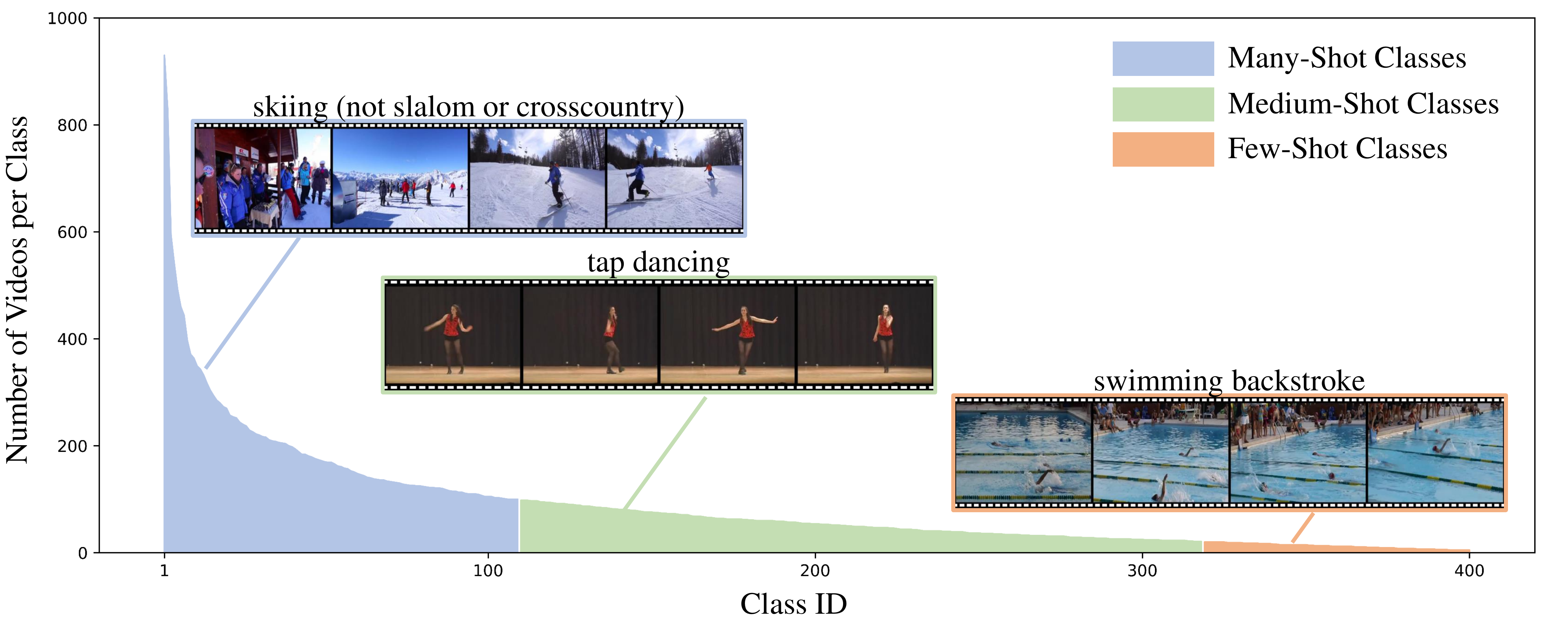}}
    \caption{\textbf{The dataset statistics of Kinetics-LT.}}
    \label{fig:lt_distribution}
    \vspace{-10pt}
\end{figure*}

\subsection{Kinetics-LT}
For the long-tailed case, we construct the Kinetics-LT dataset,
which is a long-tailed version of Kinetics400 by sampling a subset following the Pareto distribution~\cite{reed2001pareto}
similar to ImageNet-LT~\cite{liu2019large}, with 930\textasciitilde5 videos per class from the 400 classes of Kinetics400 dataset.
Videos are randomly selected based on the distribution values of each class,
and the 400 classes are randomly split into 109 many-shot classes, 209 medium-shot classes, and 82 few-shot classes.
We randomly select 20 training videos per class from the original training set as the validation set.
The original validation set of Kinetics400 is used as the testing set in this paper.
The dataset specifications are shown in Figure~\ref{fig:lt_distribution}.

\subsection{Kinetics-Fewshot}
For the few-shot case, we conduct two kinds of few-shot settings, \textit{i.e.}, \textbf{5-shot-5-way} and \textbf{5-shot-C-way}.

For the \textbf{5-shot-5-way setting}, 
we adopt the few-shot version of Kinetics~\cite{zhu2018compound,zhu2020label}, 
which has been frequently used to evaluate few-shot video recognition 
in previous works~\cite{zhu2018compound,zhu2020label,bishay2019tarn,zhang2020few,cao2020few,perrett2021temporal}.
In this setup, 100 videos from 100 classes are selected, with 64, 12 and 24 classes used for train/val/test.
We conduct 200 trials with random samplings, to ensure the statistical significance.

Specifically, the \textbf{train action categories} are sampled from:
\textit{air drumming, arm wrestling, beatboxing, biking through snow, blowing glass,
blowing out candles, bowling, breakdancing, bungee jumping, catching or throwing baseball,
cheerleading, cleaning floor, contact juggling, cooking chicken, country line dancing,
curling hair, deadlifting, doing nails, dribbling basketball, driving tractor, drop kicking,
dying hair, eating burger, feeding birds, giving or receiving award, hopscotch, jetskiing, jumping into pool,
laughing, making snowman, massaging back, mowing lawn, opening bottle, playing accordion, playing badminton,
playing basketball, playing didgeridoo, playing ice hockey, playing keyboard, playing ukulele, playing xylophone,
presenting weather forecast, punching bag, pushing cart, reading book, riding unicycle, shaking head, sharpening pencil,
shaving head, shot put, shuffling cards, slacklining, sled dog racing, snowboarding, somersaulting, squat, surfing crowd,
trapezing, using computer, washing dishes, washing hands, water skiing, waxing legs, weaving basket}.

The \textbf{val action categories} are sampled from:
\textit{baking cookies, crossing river, dunking basketball, feeding fish,
flying kite, high kick, javelin throw, playing trombone,
scuba diving, skateboarding, ski jumping, trimming or shaving beard}.

The \textbf{test action categories} are sampled from:
\textit{blasting sand, busking, cutting watermelon, dancing ballet, dancing charleston,
dancing macarena, diving cliff, filling eyebrows, folding paper, hula hooping, hurling (sport),
ice skating, paragliding, playing drums, playing monopoly, playing trumpet, pushing car,
riding elephant, shearing sheep, side kick, stretching arm, tap dancing, throwing axe, unboxing}.

For the \textbf{5-shot-C-way setting}, 
we follow~\cite{ju2021prompting} to sample 5 videos from all categories to construct the training dataset,
and measure the performance on the standard validation set, \textit{i.e.} 
all videos from all categories in the validation set of Kinetics400.
For statistical significance, we also conduct 10 random sampling rounds to choose training videos.

\subsection{Kinetics-Open}
For the open-set case, we split the Kinetics400 into two parts,
with 250 categories for training and the remaining 150 categories for evaluation.
Videos in the training set and validation set are from different categories.

Specifically, the \textbf{train action categories} are sampled from:
\textit{air drumming, answering questions, applying cream, archery, arm wrestling,
arranging flowers, assembling computer, baby waking up, balloon blowing,
bandaging, barbequing, bartending, bee keeping, belly dancing, bending back,
bending metal, biking through snow, blowing glass, blowing nose, blowing out candles,
bookbinding, bouncing on trampoline, breading or breadcrumbing, breakdancing, brush painting,
brushing teeth, bungee jumping, carrying baby, cartwheeling, carving pumpkin,
catching or throwing frisbee, celebrating, changing oil, checking tires, cheerleading,
chopping wood, clean and jerk, cleaning floor, cleaning gutters, cleaning shoes, cleaning toilet,
cleaning windows, climbing a rope, climbing ladder, climbing tree, cooking egg, cooking sausages,
counting money, cracking neck, crossing river, crying, cutting nails, cutting watermelon,
dancing charleston, decorating the christmas tree, digging, disc golfing, diving cliff,
doing laundry, doing nails, drinking, drinking beer, drinking shots, driving car, driving tractor,
drumming fingers, dunking basketball, dying hair, eating cake, eating carrots, eating chips,
eating doughnuts, eating hotdog, eating spaghetti, egg hunting, exercising with an exercise ball,
faceplanting, feeding fish, filling eyebrows, flipping pancake, folding napkins, folding paper,
front raises, frying vegetables, garbage collecting, gargling, getting a haircut, giving or receiving award,
golf chipping, golf driving, grinding meat, grooming dog, grooming horse, headbanging, headbutting,
high kick, hitting baseball, hockey stop, holding snake, hopscotch, hoverboarding, hugging, hula hooping,
hurdling, hurling (sport), ice climbing, ice skating, javelin throw, jetskiing, jogging, juggling fire,
juggling soccer ball, jumpstyle dancing, kicking soccer ball, kissing, krumping, laying bricks, making bed,
making pizza, making snowman, making sushi, making tea, massaging feet, massaging person's head, mopping floor,
motorcycling, moving furniture, opening present, parasailing, passing American football (in game), peeling apples,
petting animal (not cat), petting cat, picking fruit, planting trees, plastering, playing accordion, playing badminton,
playing bagpipes, playing basketball, playing bass guitar, playing cello, playing chess, playing clarinet,
playing cymbals, playing didgeridoo, playing drums, playing flute, playing guitar, playing harmonica, playing harp,
playing ice hockey, playing keyboard, playing monopoly, playing organ, playing paintball, playing piano,
playing squash or racquetball, playing tennis, playing trombone, playing ukulele, playing violin, playing xylophone,
pole vault, presenting weather forecast, pull ups, punching person (boxing), push up, pushing car, pushing wheelchair,
reading book, riding camel, riding mountain bike, riding mule, riding scooter, riding unicycle, rock climbing,
roller skating, running on treadmill, sailing, salsa dancing, sanding floor, scuba diving, setting table, shaking head,
shaving legs, shearing sheep, shining shoes, shooting basketball, shooting goal (soccer), shoveling snow, shredding paper,
sign language interpreting, singing, ski jumping, skiing crosscountry, skiing slalom, skipping rope, skydiving, slacklining,
snatch weight lifting, sniffing, snowboarding, somersaulting, spinning poi, spraying, springboard diving, squat, stomping grapes,
stretching leg, surfing crowd, swimming breast stroke, swimming butterfly stroke, swing dancing, swinging legs, swinging on something,
tango dancing, tap dancing, tapping guitar, tasting beer, testifying, throwing discus, tobogganing, tossing coin, training dog,
trapezing, trimming or shaving beard, triple jump, unboxing, using computer, using remote controller (not gaming), using segway,
vault, waiting in line, walking the dog, washing feet, washing hair, water skiing, watering plants, waxing eyebrows, waxing legs,
weaving basket, welding, whistling, windsurfing, wrapping present, wrestling, writing, yawning, zumba}.

The \textbf{validation action categories} are sampled from:
\textit{abseiling, applauding, auctioning, baking cookies, beatboxing, bench pressing, blasting sand,
blowing leaves, bobsledding, bowling, braiding hair, brushing hair, building cabinet, building shed,
busking, canoeing or kayaking, capoeira, catching fish, catching or throwing baseball, catching or throwing softball,
changing wheel, clapping, clay pottery making, cleaning pool, contact juggling, cooking chicken, cooking on campfire,
country line dancing, crawling baby, curling hair, cutting pineapple, dancing ballet, dancing gangnam style, dancing macarena,
deadlifting, dining, dodgeball, doing aerobics, drawing, dribbling basketball, drop kicking, eating burger, eating ice cream,
eating watermelon, exercising arm, extinguishing fire, feeding birds, feeding goats, finger snapping, fixing hair, flying kite,
folding clothes, getting a tattoo, golf putting, gymnastics tumbling, hammer throw, high jump, ice fishing, ironing, juggling balls,
jumping into pool, kicking field goal, kitesurfing, knitting, laughing, long jump, lunge, making a cake, making a sandwich,
making jewelry, marching, massaging back, massaging legs, milking cow, mowing lawn, news anchoring, opening bottle, paragliding,
parkour, passing American football (not in game), peeling potatoes, playing cards, playing controller, playing cricket,
playing kickball, playing poker, playing recorder, playing saxophone, playing trumpet, playing volleyball, pumping fist,
pumping gas, punching bag, pushing cart, reading newspaper, recording music, riding a bike, riding elephant, riding mechanical bull,
riding or walking with horse, ripping paper, robot dancing, rock scissors paper, scrambling eggs, shaking hands, sharpening knives,
sharpening pencil, shaving head, shot put, shuffling cards, side kick, situp, skateboarding, skiing (not slalom or crosscountry),
slapping, sled dog racing, smoking, smoking hookah, sneezing, snorkeling, snowkiting, snowmobiling, spray painting, sticking tongue out,
stretching arm, strumming guitar, surfing water, sweeping floor, swimming backstroke, sword fighting, tai chi, taking a shower,
tapping pen, tasting food, texting, throwing axe, throwing ball, tickling, tossing salad, trimming trees, tying bow tie,
tying knot (not on a tie), tying tie, unloading truck, washing dishes, washing hands, water sliding, waxing back, waxing chest, yoga}.

\subsection{Text Description}
The text descriptions are mainly crawled from Wikipedia~\cite{wiki} and wikiHow~\cite{wikihow}.
Following~\cite{tian2021vl}, we first use the label name as the keyword to search for the best matching entry.
Then, we filter out some irrelated parts of the entries, such as "references", "external links", and "bibliography", \textit{etc.},
to obtain the external text descriptions for each class.
In addition, we also append 96 prompt sentences for each class as basic descriptions,
which are generated by filling the pre-set templates, like `\texttt{a video of a \{label\}}', with label names.

In Figure~\ref{fig:text_vis}, we display a part of text descriptions collected for our benchmarks.
We see that it is inevitable to include some noisy text descriptions, 
since these texts are all crawled from the Internet without fine-grained cleaning.
In addition, we also report the detailed statistics of the collected text descriptions in Table~\ref{tbl:txt_statics}.
It can be seen that the text quantity of different classes varies significantly.

\section{Notations}
\label{sec:method_details}
we summarize the notations used in the paper in Table~\ref{tbl:notations}.

\begin{table}[h!]
    \centering
    \caption{\textbf{Summary of notations used in the paper.}}
    \resizebox{1.\linewidth}{!}{
    \begin{tabular}{l|l}
    \shline
    Notation & Meaning \\
    \shline
    ${\Phi}_\text{video}$ & Video encoder \\
    ${\Phi}_\text{text}$ & Language encoder \\
    ${\Phi}_\text{img}(\cdot)$ & Frame encoder \\
    ${\Phi}_\text{temp}(\cdot)$ & Temporal module \\
    $\mathcal{V}\!=\!\left\{V_i \right\}_{i\!=\!1}^N$ & A batch of $N$ video samples \\
    $\mathcal{T}\!=\!\left\{T_i \right\}_{i\!=\!1}^N$ & A batch of $N$ text samples \\
    $V\!=\!\left\{I_i \right\}_{i\!=\!1}^F$ & A video of $F$ frames \\
    ${E}_{i}^T$ & Embeddings of text $T_i$ \\
    ${E}_{i}^V$ & Embeddings of video $V_i$ \\
    $\text{sim}(\cdot,\cdot)$ & Similarity function \\
    $S$ & Cosine similarity scores produced by our model \\
    $S'$ & Cosine similarity scores produced by the frozen CLIP model \\
    $M$ & Number of sampled sentences per class \\
    $\mathcal{L}_\text{text}$ & Contrastive learning NCE losses for texts \\
    $\mathcal{L}_\text{video}$ & Contrastive learning NCE losses for videos \\
    $\mathcal{L}_\text{dist}$ & Distillation loss \\
    $\mathcal{L}_\text{pred}$ & Video-Language pretraining loss \\
    $\mathcal{L}_\text{CE}$ & CrossEntropy loss \\
    $\mathcal{L}_\text{cls}$ & Language-driven GVR finetune loss \\
    $\mathbf{y}$ & Ground truth label \\
    \shline
\end{tabular}}
    \label{tbl:notations}
\end{table}

\begin{table*}[h]
    \centering
    \caption{\textbf{Detailed statistics of the text descriptions.}
        where $N_\text{min}$, $N_\text{max}$, $N_\text{mean}$, and $N_\text{Med}$ denote
        for minimum, maximum, mean, and median number of sentences of classes respectively.
        $M_\text{min}$, $M_\text{max}$, $M_\text{mean}$, and $M_\text{Med}$ denote
        for minimum, maximum, mean, and median number of words of classes respectively.
        $L_\text{Avg}$ denotes the average number of tokens per sentence.
    }
    \begin{tabular}{l|c|c|c|c|c|c|c|c|c}
    \shline
    Datasets & $N_\text{min}$ & $N_\text{max}$ & $N_\text{mean}$ & $N_\text{Med}$ & $M_\text{min}$ & $M_\text{max}$ & $M_\text{mean}$ & $M_\text{Med}$ & $L_\text{Avg}$ \\
    \shline
    Kinetics400 & 7 & 634 & 143 & 99 & 252 & 20340 & 4011 & 2605 & 28 \\
    \shline
\end{tabular}
    \label{tbl:txt_statics}
\end{table*}

\section{Implementation Details}
\label{sec:implement_details}
\noindent\textbf{Data Pre-processing.} 
If not specified, we use the segment-based input frame sampling strategy~\cite{wang2016temporal} with 8 frames. 
During training, we follow~\cite{wang2021actionclip} to process all frames to $224\!\times\!224$ input resolution.
During inference, we resize all frames to $256\!\times\!256$ and center-crop them to $224\!\times\!224$.

\noindent\textbf{Network Architectures.}
If not specified, the video encoder adopts the pre-trained CLIP~\cite{radford2021learning} visual encoder (ViT-B/16~\cite{sharir2021image}) as our frame encoder.
For the temporal module, we use a smaller version of the transformer with 6-layers and 8-head self-attention as default. To indicate the temporal order, we also add learnable temporal positional encoding onto the frame features as input.
The language encoder also follows that of CLIP~\cite{radford2021learning}, which is a 12-layer transformer, 
and the maximum length of text tokens is set to 77 (including \texttt{[SOS]} and \texttt{[EOS]} tokens).
We initialize the frame encoder and language encoder with pretrained weights of CLIP~\cite{radford2021learning} during the first stage.

\noindent\textbf{Training Hyper-parameters.}
In our implementation, we always train the models using an AdamW~\cite{loshchilov2017decoupled} optimizer with the cosine schedule~\cite{loshchilov2016sgdr}, 
a weight decay of $5\times 10^{-2}$,
and a momentum of 0.9 for 50 epochs.
During the first stage, the size of the mini-batch is set to 16, and $\alpha$ is set to 0.5.
The initial learning rate is set to $1\times 10^{-5}$ for frame encoder and language encoder,
and set to $1\times 10^{-3}$ for the temporal module.
During the second stage, the size of the mini-batch is set to 128.
Both encoders are kept frozen, and the only trainable part is the bi-modal attention head.
The learning rate of which is set to $1\times 10^{-3}$.
The number of selected sentences per class $M$ is set to 64, and $\lambda$ is set to 50.
We conduct all experiments on 8 V100 GPUs.

\section{Ablation Study}
\label{sec:abl_study}

\begin{table}[t]
    \caption{
        \textbf{Ablation studies on Kinetics-Close.}
        "Head" denotes the classification head used in stage II,
        "Bi-M" denotes the bi-modal attention head,
        "TSR" denotes the proposed text selection ruler,
        "RAND" denotes random selection strategy,
        and "BASIC" denotes only using basic prompted sentences.
    }
    \centering
    \begin{tabular}{l|c|c|c|c|c}
    \shline
    \renewcommand{\arraystretch}{0.1}
    \multirow{2}{*}{\#} & \multirow{2}{*}{Pretrain} & CLIP & \multicolumn{2}{c|}{Fine-tuning} & \multirow{2}{*}{Top-1} \\
    \cline{4-5} 
    & & Weights & Head & Ruler & \\
    \shline
    1 & $\checkmark$ & $\checkmark$ & Bi-M & TSR & \textbf{81.8} \\
    2 & - & $\checkmark$ & Bi-M & TSR & 76.0 \\
    3 & $\checkmark$ & - & Bi-M & TSR & 32.6 \\
    4 & $\checkmark$ & $\checkmark$ & FC & TSR & 79.5 \\
    5 & $\checkmark$ & $\checkmark$ & KNN & TSR & 79.9 \\
    6 & $\checkmark$ & $\checkmark$ & Bi-M & RAND & 80.0 \\
    7 & $\checkmark$ & $\checkmark$ & Bi-M & BASIC & 78.9 \\
    \shline
\end{tabular}
    \label{tbl:ablation}
\end{table}

In order to provide a deep analysis of our proposed method,
we also conduct ablation studies on the Kinetics-Close dataset.
In these experiments, we use ViT-B/16 as the default backbone.
All other settings remain the same as Sec.~\ref{sec:implement_details} unless specifically mentioned.

\begin{table}[t!]
    \caption{
            \textbf{Ablation studies on the number of layers in Temporal Module.}
            We evaluate the Top-1 accuracy on Kinetics-Close and Kinetics-Fewshot
            by using different numbers of layers in the temporal module.
        }
        \begin{subtable}[t]{0.5\textwidth}
            \small
            \caption{\textbf{Ablation studies of Temporal Module on Kinetics-Close.}}
            \centering
            \resizebox{0.93\linewidth}{!}{
            \begin{tabular}{c|cccccc}
    \shline
    Number of layers & 0 & 1 & 2 & 4 & 6 & 8 \\
    \shline
    Top-1 Acc & 79.8 & 80.9 & 81.2 & 81.4 & \textbf{81.8} & 80.6 \\
    \shline
\end{tabular}}
    \vspace{3mm}
        \end{subtable}
        \begin{subtable}[t]{0.5\textwidth}
            \caption{\textbf{Ablation studies of Temporal Module on Kinetics-Fewshot.}}
            \centering
            \resizebox{0.93\linewidth}{!}{
            \begin{tabular}{c|cccccc}
    \shline
    Number of layers & 0 & 1 & 2 & 4 & 6 & 8 \\
    \shline
    Top-1 Acc & \textbf{96.9} & 96.5 & 95.6 & 95.8 & 96.1 & 95.2 \\
    \shline
\end{tabular}}
        \end{subtable}
        \label{tbl:num_layers}
    \end{table}
    
    \begin{table*}[t!]
        \caption{
            \textbf{Effectiveness of Distillation Loss.}
            We evaluate the performance on Kinetics-LT
            to investigate the effectiveness of distillation loss. 
        }
        \centering
        \begin{tabular}{c|c|c|c|c|c|c|c}
    \shline
    Distill Kind & Dataset & Backbone & $1-\alpha$ & Top-1 & Many & Medium & Few \\
    \shline
    - & Kinetics-LT & ResNet-50 & 0 & 57.5 & 70.8 & 57.2 & 40.8 \\
    \shline
    \multirow{3}{*}{Logits} & \multirow{3}{*}{Kinetics-LT} & \multirow{3}{*}{ResNet-50} & 0.1 & 58.4 & 71.9 & 58.2 & 40.7 \\
     &  & & 0.5 & \textbf{60.8} & 71.7 & \textbf{60.4} & \textbf{47.2} \\
     &  & & 0.9 & 54.7 & 64.4 & 54.4 & 42.5 \\
    \shline
    \multirow{3}{*}{Feature} & \multirow{3}{*}{Kinetics-LT} & \multirow{3}{*}{ResNet-50} & 0.1 & 58.1 & 71.4 & 58.1 & 40.7 \\
     & & & 0.5 & 59.7 & 72.2 & 59.6 & 43.0 \\
     & & & 0.9 & 58.2 & 71.5 & 57.6 & 41.7 \\
    \shline
\end{tabular}
        \label{tbl:dist_loss}
    \end{table*}

\subparagraph{Video-Language Pre-training.}
To examine the effectiveness of our video-language pretraining (VLP) framework,
we remove it by directly performing the finetuning process on the pretrained weights of CLIP~\cite{radford2021learning}.
As reported in the \#1 and \#2 of Table~\ref{tbl:ablation},
the model with VLP outperforms the one without VLP by 5.8 points on the top-1 accuracy.
Such a gap might be attributed to the difficulties in learning
temporal information and semantic inconsistency between videos and text representation,
which can be alleviated by our VLP.

\subparagraph{CLIP Pre-trained Weights.}
To analyze the influence of CLIP pre-trained weights,
we train our method with randomly initialized weights.
Comparing the \#1 and \#3 of Table~\ref{tbl:ablation},
we can see that initializing with CLIP pre-trained weights benefits our approach.
This phenomenon is caused by the limited text corpus for pre-training.
There are only 400 class descriptions (about 95K sentences) for Kinetics400,
and it is easy to overfit a video to a specific set of sentences without a pre-trained linguistic encoder.

\subparagraph{Bi-modal Attention Head.}
We investigate the effectiveness of bi-modal attention head by comparing it with other recognition heads,
including FC (only video-based), and KNN (video-language based).
As reported in \#1, \#4 and \#5 of Table~\ref{tbl:ablation},
the proposed head performs better than FC and KNN by 2.3\% and 1.9\% points respectively.
It is notable that, as another bi-modal head, KNN also works better than FC.
These results indicate the superiority of bi-modal attention head and the power of video-language representation.

\subparagraph{Salient Sentences.}
We study the significance of the sampled salient sentences by replacing them with those sampled by
"Random" and "Basic" strategies.
For "Random", we randomly select $M$ sentences from text descriptions.
For "Basic", we only use the basic prompt sentences as the salient sentences.
As shown in Table~\ref{tbl:ablation}, the model with TSR (See the \#1 of Table~\ref{tbl:ablation}) 
outperforms the model with other strategies on the Top-1 accuracy.
It indicates the effectiveness of our TSR to filter out some noisy sentences.

\subparagraph{Temporal Module.}
We investigate the effectiveness of the temporal module by 
using different numbers of layers in the temporal module.
As shown in Table~\ref{tbl:num_layers}, the model achieves the highest recognition accuracy on Kinetics-Close
with 6 transformer layers in the temporal module.
An interesting phenomenon is that increasing the number of layers in the temporal module leads to
a significant rise in accuracy performance at the beginning, but the accuracy falls
when the temporal module has more than 6 layers.
It may be attributed to the overfitting caused by using the transformer with too many layers.
In the few-shot setting, the model achieves the highest recognition accuracy without the additional temporal module, since there are few videos to feed the data-hungry Transformer layers in the few-shot case.

\begin{table}[t!]
    \caption{
            \textbf{Ablation studies of text descriptions.}
            "NO TEXT" denotes using no text descriptions for training, same as the CLIP-Close and CLIP-LT.
            "BASIC" denotes only using basic prompted sentences for training,
            and "FULL" denotes using both basic prompted sentences and crawled text descriptions for training.
        }
        \begin{subtable}[t]{0.5\textwidth}
            \small
            \caption{\textbf{Ablation studies of text descriptions on Kinetics-Close.}}
            \centering
            \resizebox{0.9\linewidth}{!}{
            \begin{tabular}{c|c|c|c|c}
        \shline
        Method & Backbone & Operation & Top-1 & Top-5 \\
        \shline
        CLIP-Close & \multirow{3}{*}{ViT-B/16} & NO TEXT & 78.9 & 93.5 \\
        VLG & & BASIC & 78.9 & 94.8 \\
        VLG & & FULL & \textbf{81.8} & \textbf{95.3} \\
        \shline
    \end{tabular}}
    \vspace{3mm}
        \end{subtable}
        \begin{subtable}[t]{0.5\textwidth}
            \caption{\textbf{Ablation studies of text descriptions on Kinetics-LT.}}
            \centering
            \resizebox{0.92\linewidth}{!}{
            \begin{tabular}{c|c|c|c|c|c|c}
        \shline
        Method & Backbone & Operation & Overall & Many & Medium & Few \\
        \shline
        CLIP-LT & \multirow{3}{*}{ResNet-50} & NO TEXT & 53.4 & 70.3 & 53.3 & 31.3 \\
        VLG & & BASIC & 57.8 & 71.4 & 56.9 & 35.8 \\
        VLG & & FULL & \textbf{60.8} & \textbf{71.7} & \textbf{60.4} & \textbf{47.2} \\
        \shline
    \end{tabular}}
    \vspace{-10pt}
        \end{subtable}
        \label{tab:txt_ablation}
    \end{table}

\subparagraph{Distillation Loss.}
To better investigate the effectiveness of distillation loss on
reducing the risk of overfitting caused by limited text corpus.
We conduct the ablation study on Kinetics-LT, 
which has fewer videos to avoid the influence of excessive visual information,
and use ResNet-50 as the backbone.
As shown in Table~\ref{tbl:dist_loss}, our method with distillation loss achieves
higher performance in medium-shot, few-shot and overall cases, compared to the one without distribution loss.
It indicates that the distillation loss helps the model learn better video-language representation with limited data.

To further study the influence of distillation in the pre-training stage,
we try to use the pre-trained CLIP model as the teacher model to distill the video and
language encoder of our model at the feature level, in addition to the logits distillation.
As shown in Table~\ref{tbl:dist_loss}, 
both feature distillation and logits distillation with $\alpha$ of 0.5
can improve the performance in many-shot, medium-shot, few-shot and overall cases.
And our method achieves the highest performance on Kinetics-LT 
when using logits distillation with the loss weight $\alpha$ of 0.5.

\subparagraph{Text Descriptions.}
We study the significance of our collected text descriptions by replacing them with "Using no sentences" operation and "Only using basic prompted sentences" operation, both in Kinetics-Close and Kinetics-LT. 
As shown in Table~\ref{tab:txt_ablation}, using the extra class-wise text description crawled from Wiki and Wikihow can significantly improve the performance in both Kinetics-Close and Kinetics-LT. Specifically, in the few-shot classes of Kinetics-LT, one can observe that VLG with both basic prompted sentences and crawled text description gets absolute accuracy gain against VLG with only basic prompts and VLG without text descriptions (47.2\% vs. 35.8\% vs. 31.3\%). It indicates the validity of text descriptions from the Internet, and effectiveness of leveraging abundant semantic knowledge to make up for the lack of video data.

\begin{table}[t]
    \centering
    \caption{
            \textbf{Ablation studies of loss terms.}
            We investigate the effectiveness of the two terms in Eq.4 of the main body by adopting three operations: "only $P^V$", "only $P^T$", and "both $P^V$ and $P^T$".
        }
        \begin{subtable}[t]{0.5\textwidth}
            \caption{\textbf{Ablation studies of loss terms on Kinetics-Close.}}
            \centering
            \resizebox{0.9\linewidth}{!}{
            \begin{tabular}{c|c|c|c|c}
        \shline
        Method & Backbone & Operation & Top-1 & Top-5 \\
        \shline
        \multirow{3}{*}{VLG} & \multirow{3}{*}{ViT-B/16} & Only $P^V$ & 79.5 & 94.7 \\
         & & Only $P^T$ & 80.7 & 95.1 \\
         & & $P^V$ and $P^T$ & \textbf{81.8} & \textbf{95.3} \\
        \shline
    \end{tabular}}
    \vspace{3mm}
        \end{subtable}
    
        \begin{subtable}[t]{0.5\textwidth}
            \caption{\textbf{Ablation studies of loss terms on Kinetics-LT.}}
            \centering
            \resizebox{0.92\linewidth}{!}{
            \begin{tabular}{c|c|c|c|c|c|c}
        \shline
        Method & Backbone & Operation & Overall & Many & Medium & Few \\
        \shline
        \multirow{3}{*}{VLG} & \multirow{3}{*}{ResNet-50} & Only $P^V$ & 56.9 & \textbf{73.2} & 57.1 & 34.7 \\
         & & Only $P^T$ & 59.8 & 67.2 & 60.2 & \textbf{48.8} \\
         & & $P^V$ and $P^T$ & \textbf{60.8} & 71.7 & \textbf{60.4} & 47.2 \\
        \shline
    \end{tabular}}
        \end{subtable}
        \label{tab:term_ablation}
        \vspace{-3mm}
    \end{table}

\subparagraph{Loss Terms in Stage II.}
In Eq.~\ref{eq:two_terms}, 
the first term $P^V$ is based on the video-only embedding $E^V$, and the second term $P^T$ is based on the enhanced text embedding $G$. The first term adopts the MLP to obtain the classification probability, and the second term calculates the cosine similarity between the video-only embedding $E^V$ and the enhanced text embedding $G$. To study the effectiveness of these two terms, we add experiments by adopting the first term, the second term, or both in the close set and long-tailed set.

It can be seen in Table~\ref{tab:term_ablation} that in the close set, the model with both of the two terms performs better than the others, indicating the power of video-language representation when given abundant training data. In the long-tailed case, the model with only $P^V$ performs well in the "Many" case but performs poorly in the "Few" case, while the model with only $P^T$ performs well in the "Few" case but performs poorly in the "Many" case. By contrast, the model with both of the two terms serves as the trade-off without sacrificing too much performance for all cases, and further improves the overall accuracy for the long-tailed datasets. Therefore, we hold that both the two terms are necessary.

\begin{table}[t!]
    \caption{
        \textbf{Ablation studies of different splitting strategies.}
        "RAND" denotes using the original splits in our Kinetics-LT, which are randomly chosen.
        "GOOGLE" denotes using the splits sorted by the number of entries in Google search.
    }
    \centering
    \resizebox{1.\linewidth}{!}{
    \begin{tabular}{c|c|c|c|c|c|c}
    \shline
    Method & Backbone & Operation & Overall & Many & Medium & Few \\
    \shline
    \multirow{2}{*}{VLG} & \multirow{2}{*}{ResNet-50} & RAND & 60.8 & 71.7 & 60.4 & 47.2 \\
      & & GOOGLE & 61.2 & 71.4 & 62.7 & 44.2 \\
    \shline
\end{tabular}}
    \label{tbl:resplit}
    \vspace{-10pt}
\end{table}

\section{Additional Experiments}
\label{sec:additional_exp}

\subparagraph{Re-splitting the Classes.}
To demonstrate the rationality of label splitting in our Kinetics-LT, we adopt another strategy to re-split the classes according to their number of entries in Google search. As shown in the Table~\ref{tbl:resplit}, there are no apparent changes in the recognition results, indicating the rationality of our label splitting in the long-tailed case.

\subparagraph{Experiments on UCF-101}
To further demonstrate the transferability of VLG, we also conduct experiments on UCF-101.
We compare our proposed methods with prior methods on UCF-101,
and it can be seen in Table~\ref{tbl:ucf101} that our model can achieve significant performance on this dataset.

\section{Visualization}
\label{sec:more_vis}

\subsection{Visualization of Performance}
\begin{figure}[h]
    \centering
    \resizebox{1.0\linewidth}{!}{
    \includegraphics{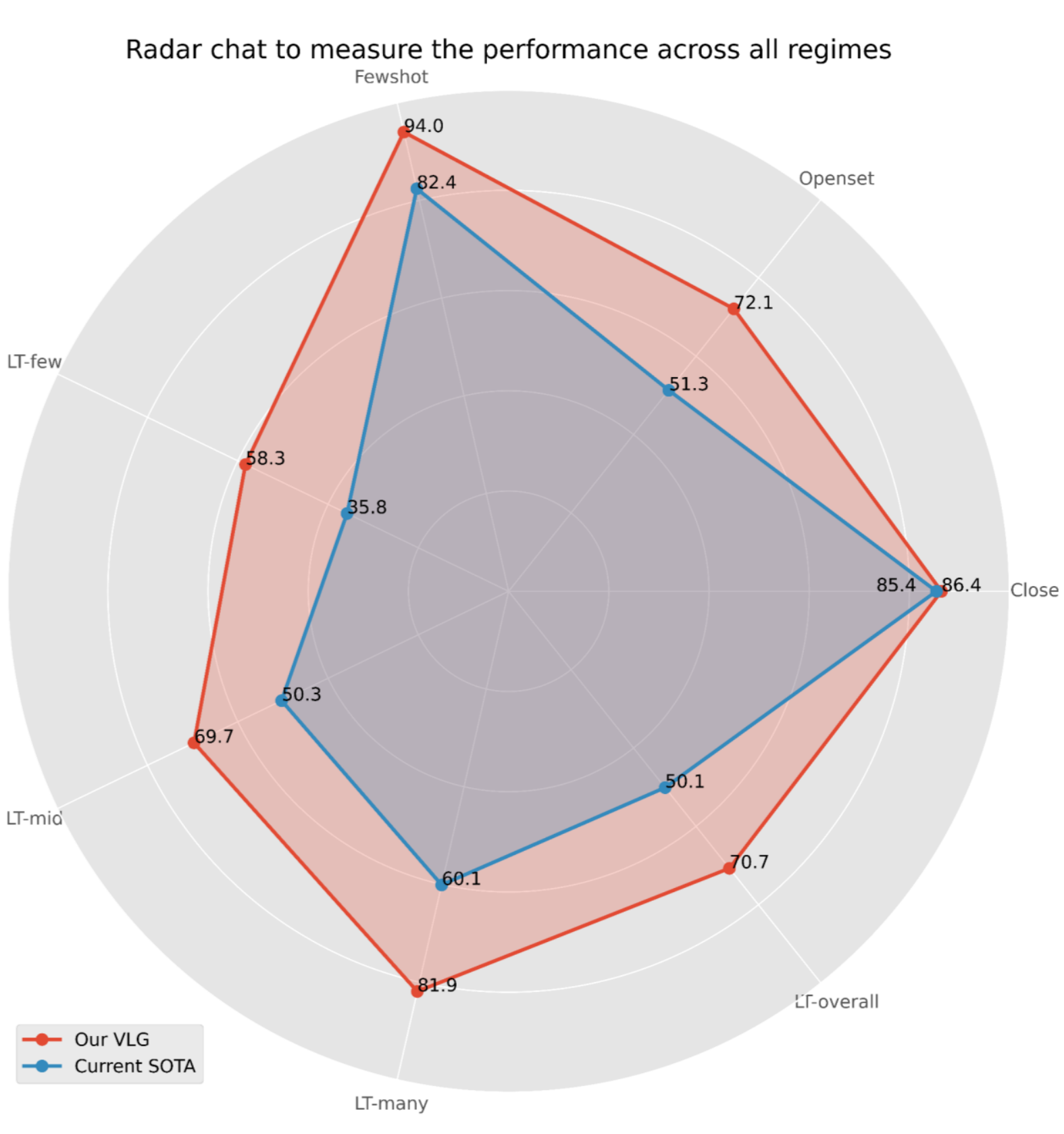}}
    \caption{\textbf{Radar chat to measure the performance across all regimes.}
        It can be seen that our method outperforms current state-of-the-art methods for all settings.
    }
    \label{fig:radar}
\end{figure}

We use a radar chart to summarize the results across all regimes in Figure~\ref{fig:radar}. The shape and area of the radar chat can serve as the total result to quantify the effectiveness and generalization ability of our method. We compare our method with current state-of-the-art methods in the radar chat, indicating the superiority of VLG over all settings.

\subsection{Class-level Performance Improvement}

\begin{figure*}[t]
    \centering
    \resizebox{1.0\linewidth}{!}{
    \includegraphics{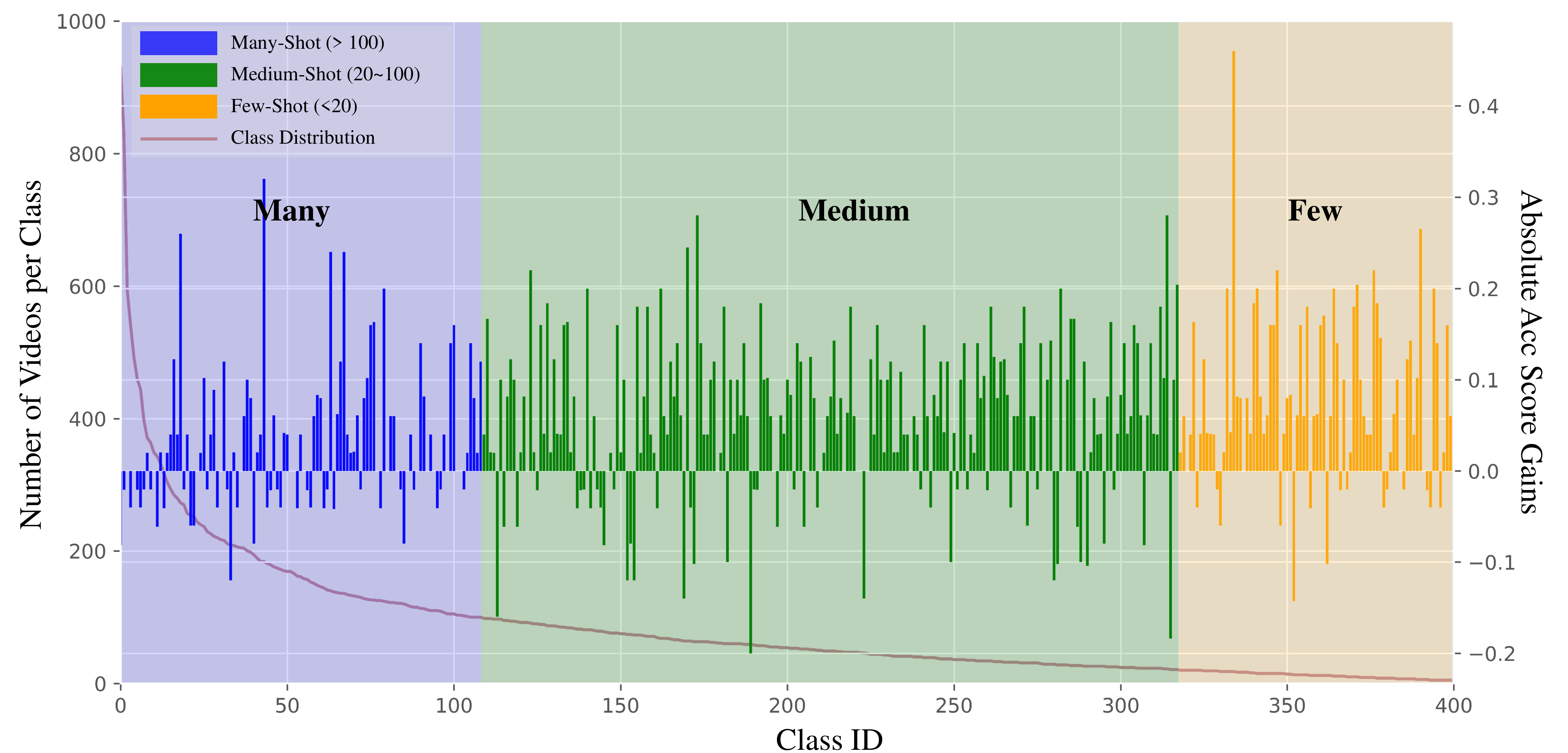}}
    \caption{\textbf{Absolute accuracy score of our method over the baseline on Kinetics-LT.}
        Our method enjoys more performance gains in classes with fewer video samples.
    }
    \label{fig:class_wise}
\end{figure*}

In Figure~\ref{fig:class_wise},
we visualize the class-level performance improvement on Kinetics-LT,
which is measured by the absolute accuracy gains of our method against the baseline,
both of which use ViT-B/16 as the visual backbone.
We observe that there are more gains in the few-shot classes,
indicating the introduced text descriptions can help mitigate the long-tailed problem.

\subsection{Visualization of Text Corpus}
\begin{figure*}[t]
    \centering
    \resizebox{0.9\linewidth}{!}{
    \includegraphics{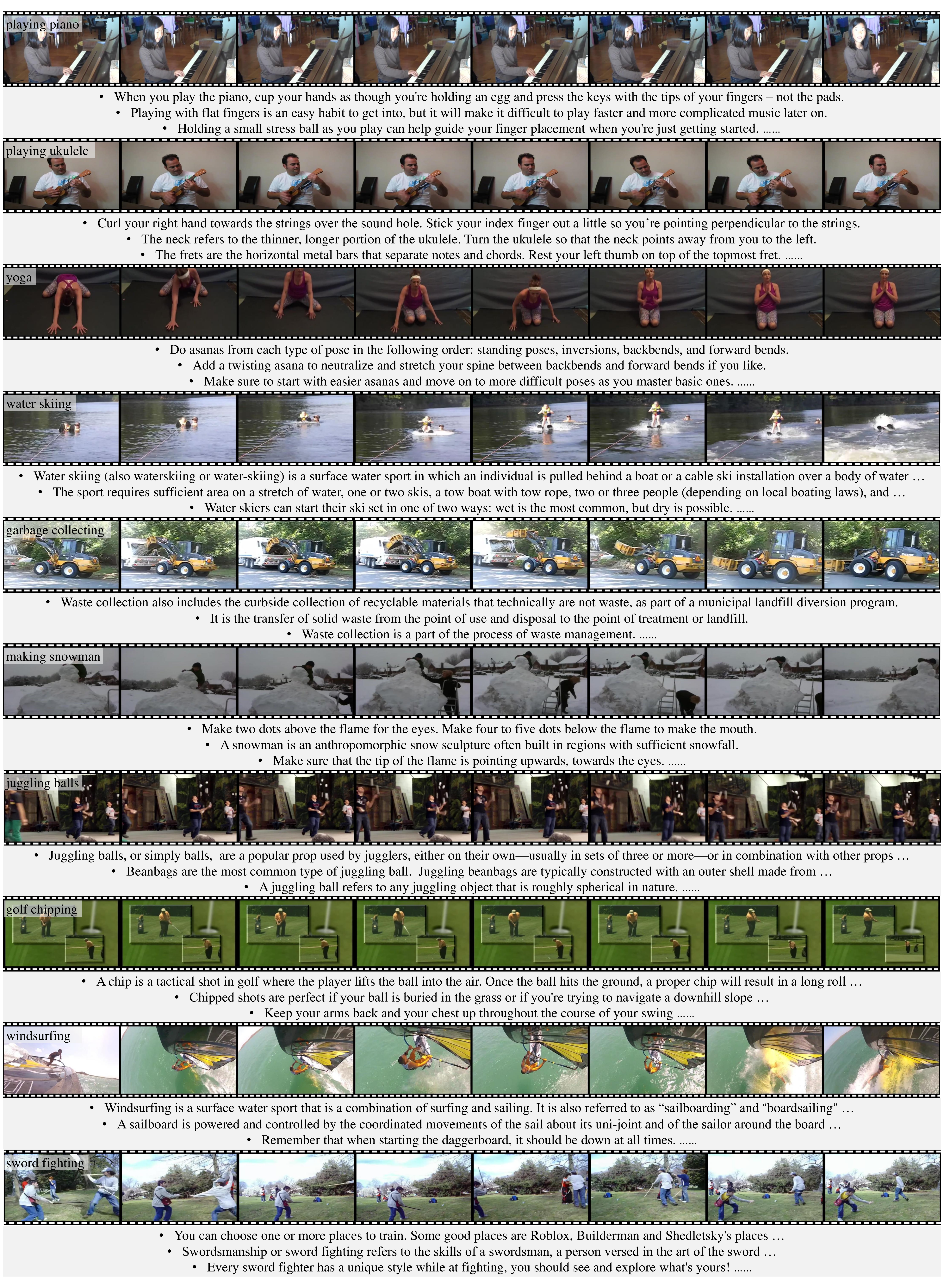}}
    \caption{\textbf{Examples of text descriptions crawled from Wikipedia and wikiHow for Kinetics400.}
        Both useful and redundant information can be found in these text corpus.
    }
    \label{fig:text_vis}
\end{figure*}

In this section, we provide some visualization of the collected text corpus in Figure~\ref{fig:text_vis}.
It can be seen that these texts contain not only some noisy information within them,
but also some static characteristics, dynamic evolution, and logical definition of the corresponding categories.

\begin{table}[t]
    \caption{
        \textbf{Results on UCF-101.}
        We report the average accuracy over three splits on UCF-101,
        $\text{CLIP}^*$ denotes that the model is initialized by the CLIP~\cite{radford2021learning} weights. K400 and K600 are used to
        denote Kinetics400 and Kinetics600 respectively.
    }
    \centering
    \resizebox{1.\linewidth}{!}{
    \begin{tabular}{c|c|c|c|c|c|c}
    \shline
    Method & Pretrain & Backbone & Frame & Views & Top-1 & Top-5 \\
    \shline
    TSN~\cite{wang2016temporal} & \multirow{6}{*}{ImageNet+K400} & ResNet-50 & 16 & - & 91.1 & - \\
    STM~\cite{jiang2019stm} &  & ResNet-50 & 16 & - & 96.2 & - \\
    S3D-G~\cite{xie2018rethinking} &  & - & - & - & 96.8 & - \\
    FASTER32~\cite{zhu2020faster} &  & - & 32 & $8 \times 1$ & 96.9 & - \\
    STAM-32~\cite{sharir2021image} & & ViT-B/16 & 32 & - & 97.0 & - \\
    R(2+1)D~\cite{tran2018closer} &  & - & - & - & 97.3 & - \\
    \shline
    D3D~\cite{stroud2020d3d} & ImageNet+K600 & ResNet-101-NL & 50 & - & 97.1 & - \\ 
    \shline
    \multirow{3}{*}{VLG} & \multirow{3}{*}{$\text{CLIP}^*$+K400} & \multirow{3}{*}{ViT-B/16} & 8 & $1 \times 1$ & 96.2 & 99.6 \\
     & & & 16 & $1 \times 1$ & 96.3 & 99.7 \\
     & & & 16 & $4 \times 3$ & 96.5 & 99.7 \\
    \shline
    \multirow{3}{*}{VLG} & \multirow{3}{*}{$\text{CLIP}^*$+K400} & \multirow{3}{*}{ViT-L/14} & 8 & $1 \times 1$ & 97.3 & 99.8 \\
    & & & 16 & $1 \times 1$ & 97.6 & 99.9 \\
    & & & 16 & $4 \times 3$ & \textbf{97.7} & \textbf{99.9} \\
    \shline
\end{tabular}
}
    \label{tbl:ucf101}
\end{table}

\subsection{More Examples of Salient Sentence}
\begin{figure*}[t]
    \centering
    \resizebox{0.9\linewidth}{!}{
    \includegraphics{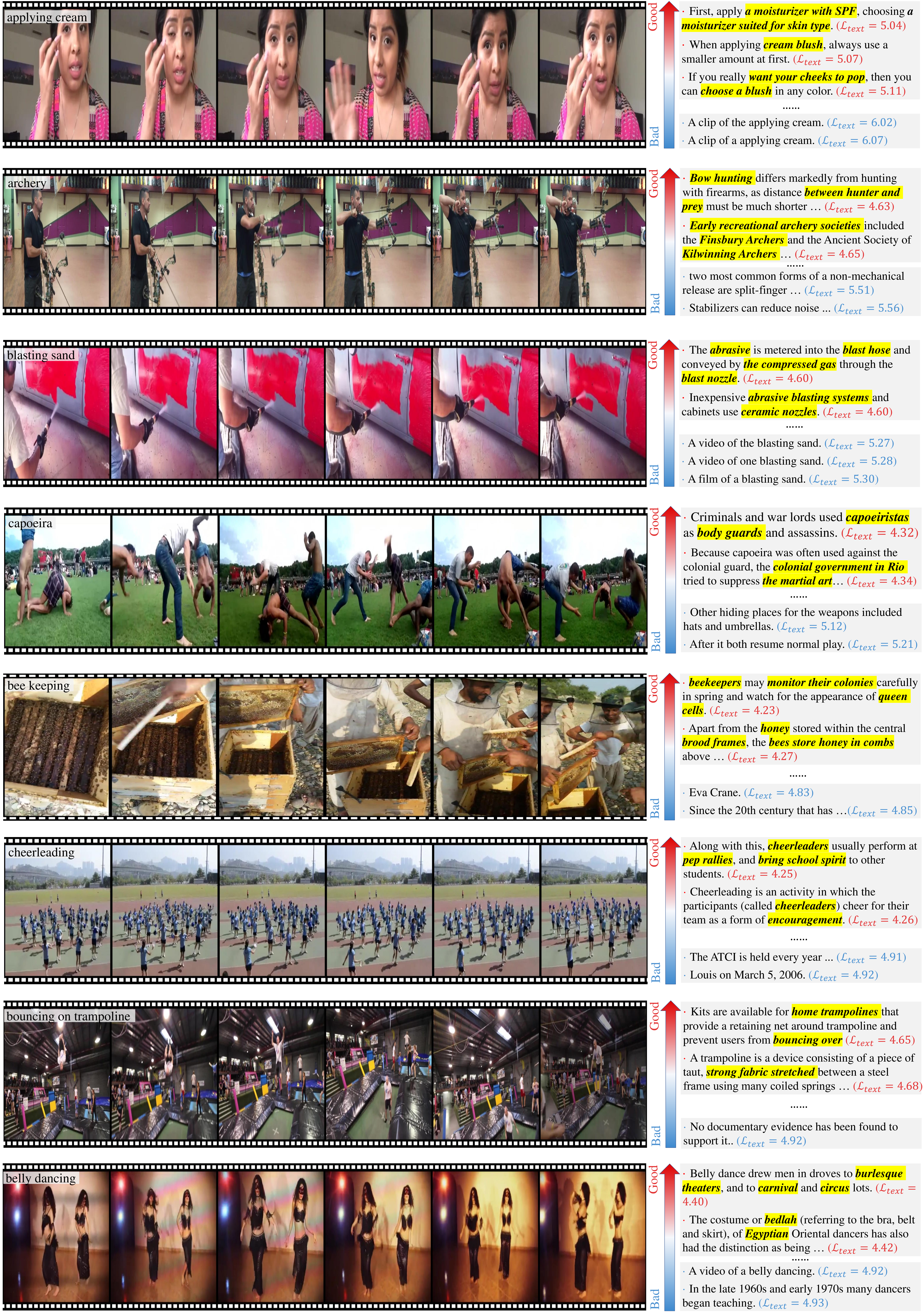}}
    \caption{\textbf{More visualization of text descriptions with corresponding $\mathcal{L}_\text{text}$.}
    The values of $\mathcal{L}_\text{text}$ reflect the saliency of these sentences, indicating the effectiveness of our proposed TSR.
    }
    \label{fig:tsr_more}
\end{figure*}

To intuitively demonstrate the effectiveness of our text selection ruler (TSR),
we provide more sentences reserved or dropped by our TSR of different categories in Figure~\ref{fig:tsr_more}.
We observe that our method can sample useful texts or filter out the useless ones.

\section{Limitation}
\label{sec:limitation}
Although our VLG achieves superior performance on multiple general video recognition settings, it still needs a two-stage training paradigm and cannot be end-to-end trained. To tackle this, we can apply reinforcement learning with reward functions~\cite{lin2022ocsampler,meng2020ar} or gumbel-softmax tricks~\cite{jang2016categorical} to further improve the non-differentiable text selection parts. 
In addition, it might be difficult to crawl suitable descriptions of labels from Wiki or WikiHow for subtle actions, like "Put the glass on top of the table". Probably, it needs to participle phrases and crawl definitions from some dictionary websites as supplementary to improve the text descriptions.

\section{Ethic Statement}
\label{sec:negative}
We use open datasets in our experiments following their licensing requirements.
Our models may be subject to biases and other possible undesired mistakes, depending on how they are trained in reality. We didn’t focus on potential negative impacts, because this work was not mainly designed for applications with potential negative impacts. As a recognition framework, it may be used for any related applications, just similar to many other general methods. But with proper usage, the proposed method could be beneficial to society.

\section{Reproducibility Statement}
\label{sec:reproduce}
We report the necessary details to reproduce the experimental results in Sec.~\ref{sec:implement_details}, including network architectures, training hyperparameters, and dataset processing steps. We will also release our code in the future.

\clearpage

\end{document}